\documentclass[10pt,twocolumn,letterpaper]{article}
\usepackage[pagenumbers]{cvpr}
\usepackage{cuted}
\usepackage{colortbl}
\usepackage{stfloats}
\usepackage{flushend}
\usepackage{microtype}

\definecolor{cvprblue}{rgb}{0.21,0.49,0.74}
\usepackage[breaklinks,colorlinks,allcolors=cvprblue,bookmarksnumbered]{hyperref}
\hypersetup{
  pdftitle={MoCA4D: Monocular 3D Bounding Box Lifting and Tracking with Camera and Object Motion},
  pdfauthor={Changwoo Jeon, Jongyoo Kim}
}

\title{MoCA4D: Monocular 3D Bounding Box Lifting and Tracking\\
with Camera and Object Motion}
\author{Changwoo Jeon\qquad Jongyoo Kim\\[3pt]Yonsei University}
\date{}

\begin{document}
\raggedbottom
\maketitle
\begin{strip}
    \centering
    \includegraphics[width=\textwidth]{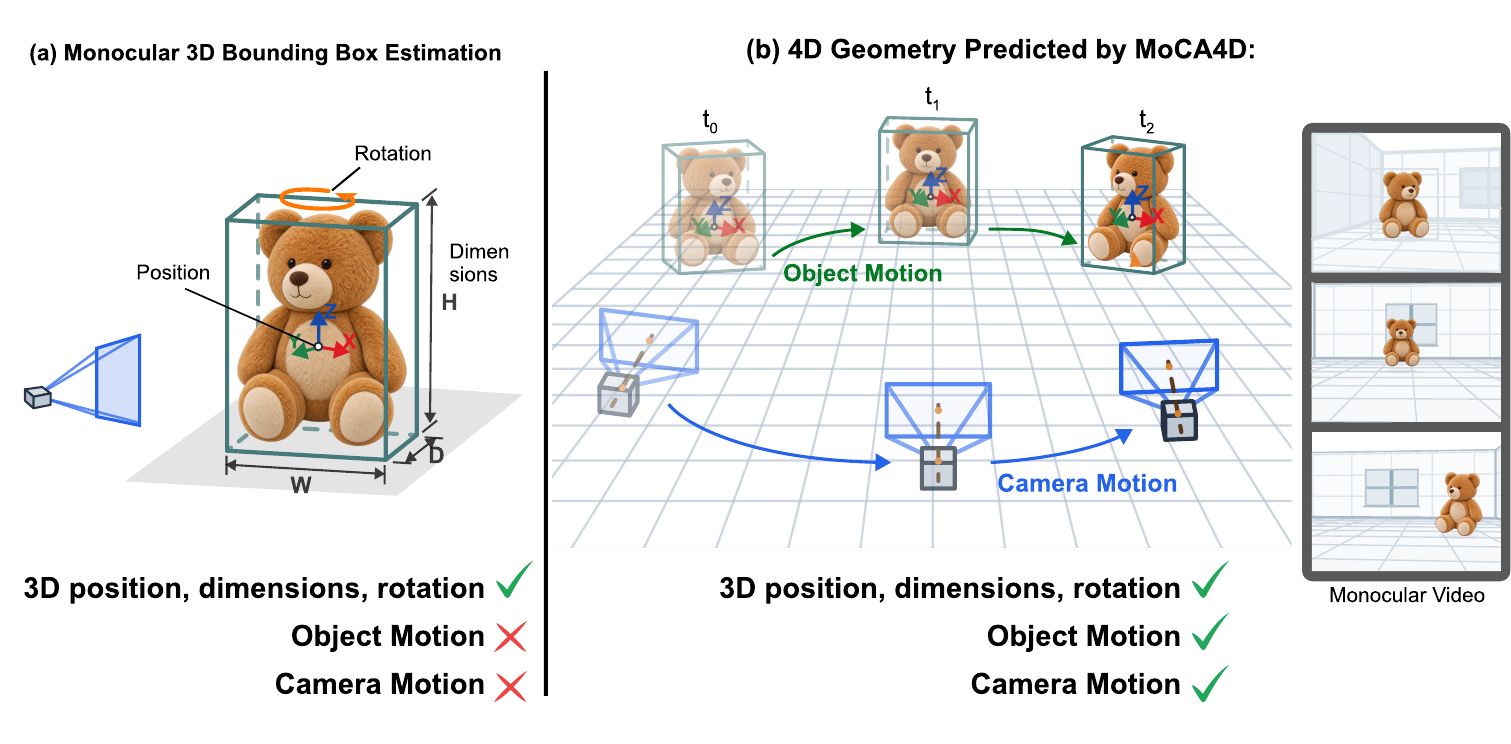}
    \captionsetup{hypcap=false}
    \captionof{figure}{\textbf{Object-centric 4D geometry predicted by MoCA4D.} MoCA4D predicts camera and object motion alongside 3D bounding box tracks from monocular video. The box tracks describe object position, orientation, and dimensions over time.}
    \label{fig:4d-geometry}
    \captionsetup{hypcap=true}
\end{strip}
\begin{abstract}
Monocular 3D bounding box tracking must account for both object motion and changes in camera viewpoint. Framewise box lifting with temporal filtering leaves these motion components implicit, while static-scene fusion assumes stationary objects. We introduce \textbf{MoCA4D}, a framework for 3D box lifting and tracking from monocular video, associated 2D box prompts, and camera intrinsics. An observation branch estimates per-frame geometry, while a dynamic branch predicts a camera transformation shared across objects and a separate rigid transformation for each object. We propose \textbf{Star Optimization} to refine current depth and orientation using the observation and historical boxes propagated through these transformations. The refined box then guides subsequent propagation. Across all evaluated objects in Argoverse 2 and ADT, Star Optimization reduces center motion error by 38.1\% and 38.7\%, respectively, relative to Kalman filtering of the same observations. On MOVi-F, the dynamic branch achieves the lowest corner and rotation errors and the highest IoU among the evaluated methods over eight-transition motion windows. Real-data camera-only controls further link predicted object motion to improved localization on dynamic-object tracks.
\end{abstract}

\section{Introduction}
\label{sec:intro}

As illustrated in Fig.~\ref{fig:4d-geometry}, MoCA4D estimates object-centric 4D geometry from monocular video, predicting 3D position, orientation, and dimensions over time, alongside object and camera SE(3) motions. These quantities support robotic manipulation~\cite{ponimatkin20256d, hsu2025spot, zhu2026vision}, dynamic-scene simulation for autonomous driving~\cite{hess2025splatad, ljungbergh2024neuroncap}, and viewpoint and object-motion control in video generation~\cite{shuai2025free, fu20253dtrajmaster}. Box trajectories also provide compact controls for character animation~\cite{yuan2026dancingbox, bahmani2024tc4d}.

Monocular image 3D detectors~\cite{jeon2026moca3d, zhang2025detect, yao2026open, brazil2023omni3d, huang2026wilddet3d} estimate per-frame object geometry without explicitly modeling motion across frames. Their tracking extensions, such as WildDet3D~\cite{huang2026wilddet3d}, apply Kalman filtering and rotation smoothing to framewise boxes without separately estimating camera and object motion. Video-based methods~\cite{hou2026detany4d, boxer2026} integrate observations across time to estimate temporally consistent 3D geometry. However, their temporal formulations are built on the assumption that objects remain static in the world frame.

MoCA4D operates in per-frame camera coordinates, estimating 3D bounding boxes and separate camera and object SE(3) motions through two branches sharing a visual backbone. The observation branch uses a box-conditioned Dense Prediction Transformer (DPT)~\cite{ranftl2021vision} and a regression head to predict geometry from a 2D prompt. The dynamic branch processes consecutive frames with correspondence features and object-conditioned spatial tokens. Dedicated camera and object queries aggregate information from these features through attention. Separate regression heads then map the updated camera queries to a shared camera SE(3) transformation and the updated object queries to object-specific SE(3) transformations.

We propose \textbf{Star Optimization} to connect the two branches through motion-guided temporal refinement. It propagates historical boxes into the current camera frame and combines them with the current observation through closed-form depth and orientation updates. The refined box then feeds back into subsequent propagation.

Our contributions are as follows:
\begin{itemize}
    \item We introduce \textbf{MoCA4D}, a monocular 3D box lifting and tracking framework that jointly predicts object geometry, shared camera motion, and object-specific rigid motion.
    \item We propose \textbf{Star Optimization}, a closed-form refinement of current depth and orientation using the observation and motion-propagated historical boxes, with recursive reuse of the refined state.
    \item MoCA4D achieves efficient box lifting and accurate rigid-motion propagation, while Star Optimization reduces tracking motion errors relative to filtering the same observations. Controlled ablations examine motion decomposition, supervision, and temporal refinement.
\end{itemize}

\section{Related Work}
\label{sec:related}

\paragraph{3D Geometry Prediction.}
3D bounding box prediction estimates an object's position, orientation, and dimensions, providing a compact representation of its geometry. Cube R-CNN~\cite{brazil2023omni3d} and OVMono3D~\cite{yao2026open} recover these parameters from object region features, while DetAny3D~\cite{zhang2025detect} uses a prompt-conditioned transformer.
MoCA3D~\cite{jeon2026moca3d} formulates box prediction in the image plane, estimating projected corners and their depths through dense heatmaps and depth maps. Other representations include normalized object coordinates~\cite{wang2019normalized} and meshes~\cite{gkioxari2019mesh}, which encode canonical surface coordinates and object shape, respectively.
MoCA4D adopts 3D bounding boxes in its observation branch, providing competitive geometric accuracy and efficient inference for subsequent motion estimation and tracking.

\paragraph{Object Motion Estimation.}
Rigid object motion can be recovered from 3D point trajectories or estimated through 6DoF pose tracking. Point-tracking methods~\cite{tapip3d, badki2026l4p} and 4D reconstruction models~\cite{luo20264rc, Sucar_2026_CVPR, jeon2026point4d, fradlin2026arrow} recover temporal 3D correspondences, from which object transformation can be obtained through fitting. SM4RT~\cite{lin2026sm4rt} explicitly models structured scene motion using rigid motion bases. For 6DoF tracking, FoundationPose~\cite{wen2024foundationpose} uses RGB-D observations with a CAD model, while BundleSDF~\cite{wen2023bundlesdf} jointly tracks and reconstructs unknown objects from RGB-D video. MoCA4D directly predicts camera and object SE(3) transformations from monocular video. This object-level parameterization preserves cuboid rigidity during propagation.

\begin{figure*}[t]
  \centering
   \includegraphics[width=\textwidth]{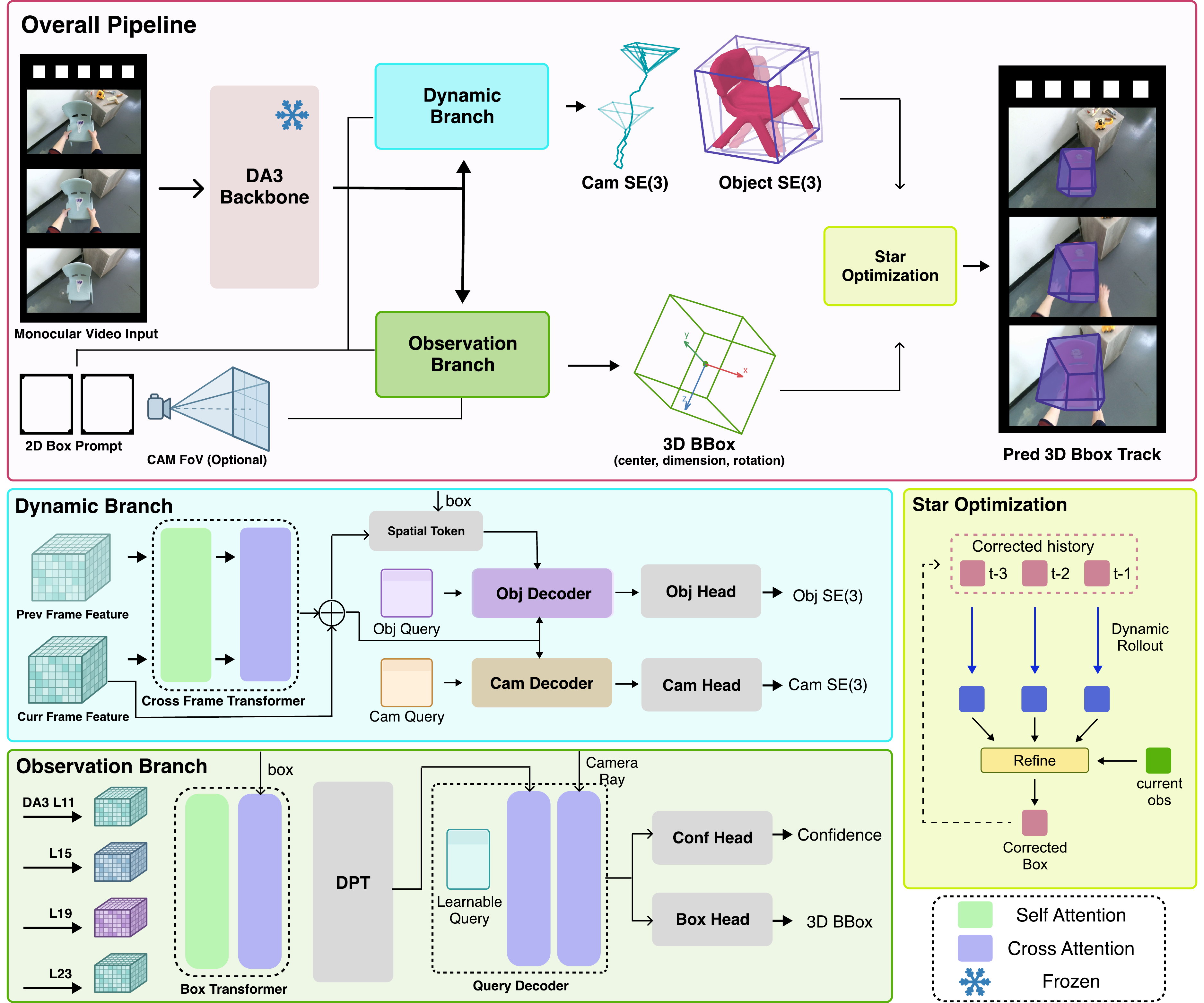}
   \caption{\textbf{MoCA4D pipeline.} A shared frozen backbone supports per-frame 3D box lifting and separate camera and object motion prediction. Star Optimization propagates corrected historical boxes into the current frame, refines the observed depth and orientation, and stores the refined box for subsequent propagation. The lower panels detail the two branches and this feedback loop.}
   \label{fig:pipeline}
\end{figure*}

\paragraph{3D Bounding Box Tracking.}
DetAny4D~\cite{hou2026detany4d} jointly estimates 3D boxes and camera poses and promotes temporal consistency in a shared world coordinate system. Boxer~\cite{boxer2026} lifts 2D prompts to 3D boxes and integrates them through tracking and fusion for static scene reconstruction. WildDet3D~\cite{huang2026wilddet3d} filters framewise boxes with Kalman filtering~\cite{kalman1960new} and quaternion smoothing. MoCA4D explicitly predicts object-specific rigid motion separately from the camera transformation, using their composition to propagate historical boxes before refining the current estimate.

\section{Method}
\label{sec:method}

\subsection{Formulation and Overall Pipeline}
\label{sec:3-1}

Given a monocular RGB sequence $\{I_t\}_{t=1}^{T}$, associated 2D box prompts $\{b_t^i\}$, and optional camera calibration, MoCA4D estimates 3D box trajectories with separate camera and object motion. Here, $t$ indexes frames, $i$ identifies objects, and $\mathbf{K}_t$ denotes the camera intrinsic matrix for frame $t$, obtained from supplied calibration or constructed from the field-of-view (FoV) predicted by DA3.

We represent a 3D bounding box as $\mathcal{B}_t^i = (\mathbf{c}_t^i, \mathbf{R}_t^i, \mathbf{s}_t^i)$, where $\mathbf{c}_t^i\in\mathbb{R}^3$ is its center, $\mathbf{R}_t^i\in\mathrm{SO}(3)$ is its orientation, and $\mathbf{s}_t^i\in\mathbb{R}_{+}^{3}$ contains its dimensions. 

Let $\mathbf{C}_t \in \mathrm{SE}(3)$ denote the camera-to-world pose at time $t$, and $\mathbf{O}_t^i \in\mathrm{SE}(3)$ the object-to-world pose of object $i$.
For a rigid object, eliminating its fixed local point coordinates gives:
\begin{equation}
\mathbf{x}_t^i=\mathbf{C}_t^{-1}\mathbf{O}_t^i(\mathbf{O}_{t-1}^i)^{-1}\mathbf{C}_{t-1}\mathbf{x}_{t-1}^i.
\label{eq:interframe-transform}
\end{equation}
Define the shared camera coordinate change and the object motion in the previous camera frame as:
\begin{equation}
\begin{aligned}
\mathbf{A}_t
&=
\mathbf{C}_t^{-1}\mathbf{C}_{t-1}, \\
\mathbf{B}_t^i
&=
\mathbf{C}_{t-1}^{-1}
\mathbf{O}_t^i
(\mathbf{O}_{t-1}^i)^{-1}
\mathbf{C}_{t-1}.
\end{aligned}
\label{eq:motion-factorization}
\end{equation}
Here, $\mathbf{B}_t^i$ applies object motion within
the previous camera coordinate frame, and
$\mathbf{A}_t$ subsequently maps the result into
the current camera coordinate frame.
Eq.~\eqref{eq:interframe-transform} therefore becomes:
\begin{equation}
\mathbf{x}_t^i
=
\mathbf{A}_t\mathbf{B}_t^i\mathbf{x}_{t-1}^i.
\label{eq:motion-propagation}
\end{equation}
The camera transformation $\mathbf{A}_t$ is shared across objects.

MoCA4D implements this formulation through observation and dynamic branches built on a shared geometric backbone, Depth Anything 3~\cite{depthanything3}. The observation branch estimates 3D bounding boxes from the current image and 2D box prompts, while the dynamic branch predicts $\hat{\mathbf{A}}_t$ and $\hat{\mathbf{B}}_t^i$ from consecutive frames. 
These transformations propagate historical box estimates into the current camera frame.
Star Optimization combines the propagated estimates with the current observation to produce the refined box $\hat{\mathcal{B}}_t^i$, whose center and orientation are reused for subsequent propagation.

\subsection{Observation Branch}
The observation branch estimates a 3D bounding box for each object in the current frame. As shown in Fig.~\ref{fig:pipeline}, we extract features from four intermediate layers of the DA3 backbone~\cite{depthanything3} and fuse them using a box-conditioned DPT~\cite{ranftl2021vision}.
Following the box-conditioning design of MoCA3D~\cite{jeon2026moca3d}, we inject a spatial prior encoding positions relative to the 2D box and apply cross-attention~\cite{vaswani2017attention} between image features and box embeddings.

We adopt a query decoder inspired by WildDet3D~\cite{huang2026wilddet3d}. For each box prompt, a learnable query attends to the encoded prompt and the fused DPT features. The query is further conditioned on camera-ray embeddings computed from $\mathbf{K}_t$, together with features from the final DA3 layer.

The resulting query feeds separate box and confidence heads.
The box head predicts the projected center as an offset from the prompt center, depth, dimensions, and a 6D rotation representation.
Given the predicted projected center
$(\hat{u}_t^i,\hat{v}_t^i)$ and depth $\hat{z}_t^i$,
the 3D center is recovered as:
\begin{equation}
\hat{\mathbf{c}}_t^i
=
\hat{z}_t^i\mathbf{K}_t^{-1}
\begin{bmatrix}
\hat{u}_t^i & \hat{v}_t^i & 1
\end{bmatrix}^{\top}.
\end{equation}
Together, these predictions define the observation box
$\hat{\mathcal{B}}_t^{i,\mathrm{obs}}$ in camera coordinates.

We supervise box parameters with $\ell_1$ losses and confidence with quality-target binary cross-entropy at every decoder layer. Adaptation accounts for the 24 equivalent cuboid parameterizations. Inference uses the final decoder output; confidence does not determine the fixed Star fusion weights. Appendix~\ref{sec:supp-observation} details the observation decoder and its losses.

\subsection{Dynamic Branch}

The dynamic branch extracts motion cues from consecutive DA3 feature maps through a cross-frame transformer that alternates within-frame self-attention and bidirectional cross-attention. Soft correspondences following GMFlow~\cite{xu2022gmflow} provide differentiable displacement cues, which are combined with appearance features through a residual MLP. Applying $7\times7$ RoI Align within the two box prompts yields 98 spatial tokens $\mathbf{Z}_t^i$, augmented with geometric and temporal embeddings.

Camera queries and auxiliary register tokens attend to global features without box conditioning.
Object queries are conditioned on both box prompts and the previous 3D center, and attend to features augmented with a spatial box prior. 
At each decoder layer, an additional cross-attention operation reads $\mathbf{Z}_t^i$ into the object queries through a learned scalar gate.
Query attention allows object queries to read global context while preventing camera and register queries from reading object queries. The camera estimate therefore remains independent of the target box and is shared across objects.

Separate prediction heads decode the camera and object
queries into 6D rotation representations and 3D motion vectors.
The rotations are converted to
$\hat{\mathbf{R}}_t^A$ and $\hat{\mathbf{R}}_t^{B,i}$.
The camera head directly predicts translation
$\hat{\mathbf{t}}_t^A$, while the object head predicts
displacement $\hat{\mathbf{d}}_t^i$ at the previous center
$\bar{\mathbf{c}}_{t-1}^i$ in the previous camera frame.
The corresponding object translation is:
\begin{equation}
\hat{\mathbf{t}}_t^{B,i}
=
\hat{\mathbf{d}}_t^i
+
\left(\mathbf{I}-\hat{\mathbf{R}}_t^{B,i}\right)
\bar{\mathbf{c}}_{t-1}^i.
\label{eq:object-pivot}
\end{equation}
The resulting transformations $\hat{\mathbf{A}}_t\hat{\mathbf{B}}_t^i$ propagate the cuboid according to Eq.~\eqref{eq:motion-propagation}.

We supervise both heads using the ground-truth transformations from Eq.~\eqref{eq:motion-factorization}.
The camera loss $\mathcal{L}_A$ combines rotation geodesic error with a Smooth L1 translation loss.
The object loss $\mathcal{L}_B$ applies Smooth L1 penalties to the rotation geodesic error and the displacement error.
The displacement target is evaluated at the same supplied center.

We additionally supervise cuboid corners propagated through successive predicted transformations and apply an auxiliary 2D endpoint-error loss to the matching displacement.
The complete objective is:
\begin{equation}
\mathcal{L}_{\mathrm{dyn}}
=
\mathcal{L}_A+\mathcal{L}_B
+0.1\,\mathcal{L}_{\mathrm{roll}}
+\mathcal{L}_{\mathrm{flow}},
\label{eq:dynamic-loss}
\end{equation}
where $\mathcal{L}_{\mathrm{roll}}$ is a Smooth L1 corner
loss over rollout windows of up to eight transitions.
The flow loss averages endpoint errors over valid matches
in normalized image coordinates and is applied where
ground-truth optical flow is available.

\begin{figure*}[t]
\centering
\includegraphics[width=\textwidth]{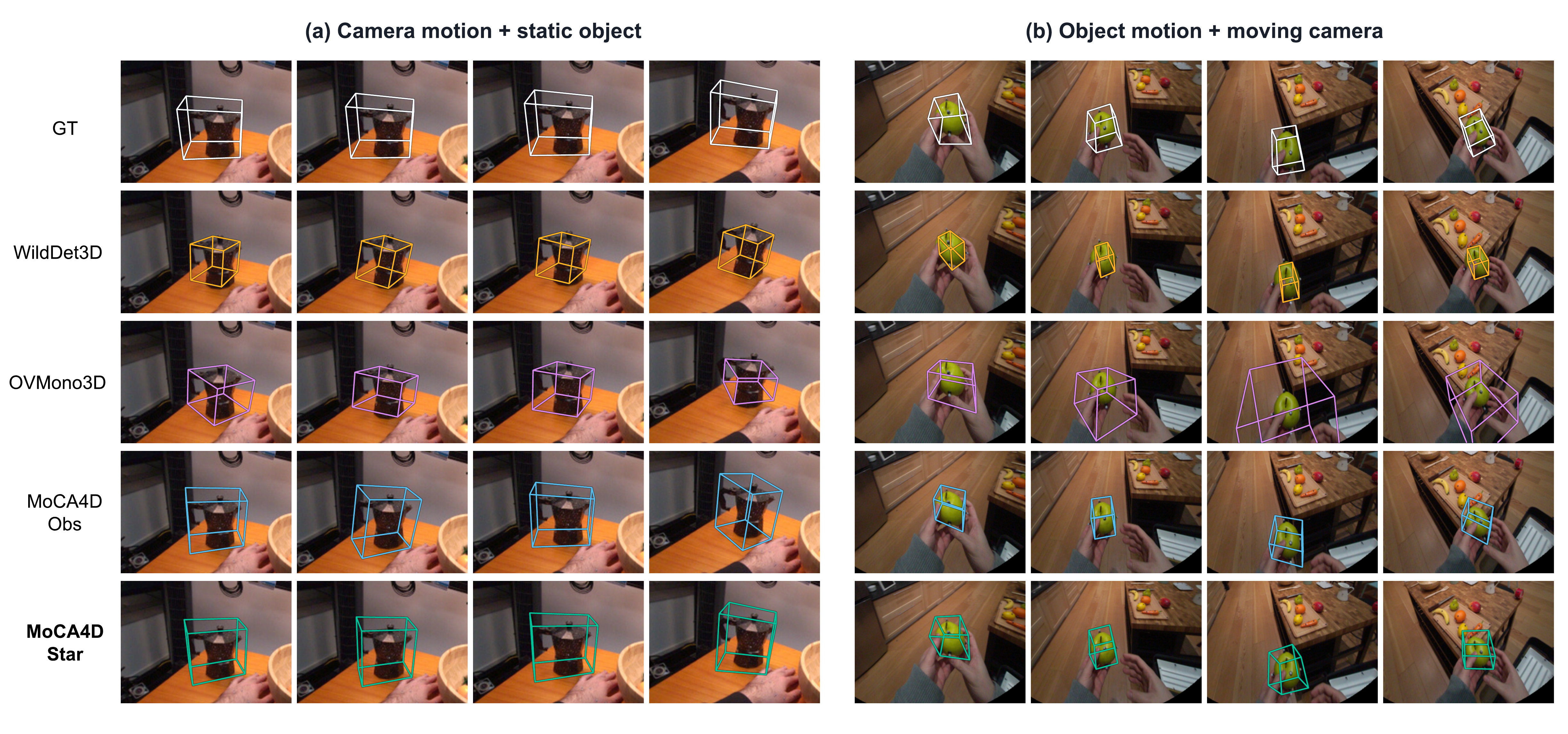}
\caption{\textbf{Qualitative comparison of monocular 3D box tracking.} Projected 3D boxes for (a) a static object viewed by a moving camera and (b) a moving object viewed by a moving camera. Compared with framewise predictions, Star uses motion-propagated history to stabilize box orientation while following changes in viewpoint and object pose. Rows show ground truth, WildDet3D, OVMono3D, MoCA4D observations, and MoCA4D with Star Optimization.}
\label{fig:qual}
\end{figure*}

\subsection{Star Optimization}
Star Optimization computes the closed-form minimizers of the fixed candidate objectives below, combining the current observation with past corrected boxes propagated by the dynamic branch.
For clarity, we omit the object index $i$ and write the current observation as
$\hat{\mathcal{B}}_t^{\mathrm{obs}}
=(\hat{\mathbf{c}}_t^{\mathrm{obs}},
\hat{\mathbf{R}}_t^{\mathrm{obs}},
\hat{\mathbf{s}}_t^{\mathrm{obs}})$,
where the three components are the 3D center, orientation, and dimensions in the current camera frame.

Let $\mathcal{H}_t$ contain the three preceding observation times. For each $\tau\in\mathcal{H}_t$, we propagate the corrected box $\hat{\mathcal{B}}_\tau$ to the time $t$ using Eq.~\eqref{eq:motion-propagation}.
Each candidate supplies its own propagated center to the dynamic branch at every step.

We denote its resulting center and orientation by $\tilde{\mathbf{c}}_{\tau\rightarrow t}$ and $\tilde{\mathbf{R}}_{\tau\rightarrow t}$.
Before combining orientations, we align equivalent cuboid axis assignments to the current observation by matching dimensions and resolving ties through rotation agreement.

We keep the observed projected center and dimensions fixed and optimize the depth and orientation.
Let $\hat{z}_t^\mathrm{obs}$ be the third component of $\hat{\mathbf{c}}_t^\mathrm{obs}$.
The center is parameterized as:
\begin{equation}
\mathbf{c}_t(z)=z\mathbf{r}_t,
\qquad
\mathbf{r}_t=
\frac{\hat{\mathbf{c}}_t^{\mathrm{obs}}}
     {\hat{z}_t^{\mathrm{obs}}}.
\label{eq:star-ray}
\end{equation}
$\mathbf{r}_t$ is the viewing ray through the observed projected center. Varying $z$ therefore changes depth while preserving the image projection.

Each propagated candidate constrains the current box, forming a star centered at time $t$.
With these candidates fixed, we optimize current depth $z$ and orientation $\mathbf{R}\in\mathrm{SO}(3)$ by minimizing:
\begin{equation}
\begin{aligned}
&E_c(z)
=
w_o
\|\mathbf{c}_t(z)
-\hat{\mathbf{c}}_t^{\mathrm{obs}}\|_2^2
\\
&\qquad
+\lambda_c
\sum_{\tau\in\mathcal{H}_t}
w_{\tau,t}
\|\mathbf{c}_t(z)
-\tilde{\mathbf{c}}_{\tau\rightarrow t}\|_2^2,
\\[3pt]
&E_R(\mathbf{R})
=
w_o
d_F(\mathbf{R},
\hat{\mathbf{R}}_t^{\mathrm{obs}})^2
\\
&\qquad
+\frac{\lambda_R}{|\mathcal{H}_t|}
\sum_{\tau\in\mathcal{H}_t}
d_F(\mathbf{R},
\tilde{\mathbf{R}}_{\tau\rightarrow t})^2,
\end{aligned}
\label{eq:star-objectives}
\end{equation}
where $d_F(\mathbf{U},\mathbf{V})=\|\mathbf{U}-\mathbf{V}\|_F/\sqrt{2}$ is the chordal rotation distance.
The weight $w_o$ controls the current observation terms, and $\lambda_c$ and $\lambda_R$ control the historical center and rotation terms.

The two objectives are separable and admit closed-form solutions.
Since $\sum_{\tau\in\mathcal{H}_t}w_{\tau,t}=1$, the optimal depth is:
\begin{equation}
z_t^\star =
\frac{
\mathbf{r}_t^\top
\left(
w_o\hat{\mathbf{c}}_t^{\mathrm{obs}}
+
\lambda_c
\sum_{\tau\in\mathcal{H}_t}
w_{\tau,t}\tilde{\mathbf{c}}_{\tau\rightarrow t}
\right)
}{
(w_o+\lambda_c)\|\mathbf{r}_t\|_2^2
}.
\label{eq:star-depth-solution}
\end{equation}
For rotation, we form the weighted matrix as:
\begin{equation}
\mathbf{M}_t =
w_o\hat{\mathbf{R}}_t^{\mathrm{obs}}
+
\frac{\lambda_R}{|\mathcal{H}_t|}
\sum_{\tau\in\mathcal{H}_t}
\tilde{\mathbf{R}}_{\tau\rightarrow t}.
\label{eq:star-rotation-matrix}
\end{equation}
Given its SVD $\mathbf{M}_t=\mathbf{U}_t\mathbf{\Sigma}_t\mathbf{V}_t^\top$, the optimal rotation is:
\begin{equation}
\mathbf{R}_t^\star =
\mathbf{U}_t
\operatorname{diag}
\left(
1,1,\det(\mathbf{U}_t\mathbf{V}_t^\top)
\right)
\mathbf{V}_t^\top.
\label{eq:star-rotation-solution}
\end{equation}
The final box is:
\begin{equation}
\hat{\mathcal{B}}_t =
\left(
\mathbf{c}_t(z_t^\star),
\mathbf{R}_t^\star,
\hat{\mathbf{s}}_t^{\mathrm{obs}}
\right).
\label{eq:star-output}
\end{equation}
Propagation applies rigid transformations to each historical cuboid. The output dimensions are re-estimated from the current observation and are not constrained to remain constant over time. We store this corrected box as a new history state.

\subsection{Training Datasets and Settings}
We train the two branches independently with a frozen DA3-Large~\cite{depthanything3} backbone. Observation pretraining uses Omni3D~\cite{brazil2023omni3d}; dynamic pretraining uses MOVi-F~\cite{greff2022kubric}. We adapt both on HOT3D~\cite{banerjee2025hot3d}, HOI4D~\cite{liu2022hoi4d}, AV2~\cite{wilson2021argoverse2}, and Virtual KITTI 2~\cite{cabon2020virtual}. Appendix~\ref{sec:supp-implementation} describes the feature representations, supervision, and training schedules.

Star Optimization requires no training. We set $w_o=0.25$, $\lambda_c=0.5$, and $\lambda_R=1$, with center weights $(1/7,2/7,4/7)$ from oldest to newest and uniform rotation weights.

\begin{table*}[t]
    \centering
    \small
    \caption{\textbf{Observation branch accuracy and inference efficiency.}
    All methods use the same tight 2D box prompts, including SAM2-derived boxes, and provided camera intrinsics.
    Bold and underlined values indicate the best and second-best results,
    respectively, in each ranked column.}
    \label{tab:omni3d-observation}

    \setlength{\tabcolsep}{4pt}
    \begin{tabular*}{\linewidth}{@{\extracolsep{\fill}}lrrrrrrrr@{}}
        \hline
        & \multicolumn{2}{c}{ARKitScenes}
        & \multicolumn{2}{c}{SUNRGBD}
        & \multicolumn{2}{c}{Omni3D}
        & \multicolumn{2}{c}{Efficiency} \\
        Method
        & PAG$_{uv}$ $\downarrow$ & IoU$_{\mathrm{3D}}$ $\uparrow$
        & PAG$_{uv}$ $\downarrow$ & IoU$_{\mathrm{3D}}$ $\uparrow$
        & PAG$_{uv}$ $\downarrow$ & IoU$_{\mathrm{3D}}$ $\uparrow$
        & Throughput $\uparrow$ & VRAM \\
        & (px) & (\%) & (px) & (\%) & (px) & (\%) & (obs/s) & (GiB) \\
        \hline
        DetAny3D      & 38.54 & 36.57 & 29.68 & \underline{37.13} & 16.51 & \underline{21.53} & 3.18 & 6.62 \\
        OVMono3D-Lift & \underline{30.04} & \textbf{41.43} & \underline{25.54} & \textbf{37.69} & 14.030 & \textbf{23.195} &\underline{6.71} & 7.69\\
        WildDet3D     & 30.16 & \underline{38.02} & 26.63 & 33.25 & \textbf{13.37} & 21.29 & 4.11 & \underline{6.42} \\
        \hline
        MoCA4D (ours) & \textbf{29.23} & 37.88 & \textbf{24.22} & 36.57 & \underline{13.790} & 19.874 & \textbf{11.52} & \textbf{1.65} \\
        \hline
    \end{tabular*}
\end{table*}

\begin{table*}[t]
\centering\small
\caption{\textbf{Rigid object motion on MOVi-F with GT reset-8.} Cuboids are reinitialized from ground truth every eight frame transitions. Offline variants access future frames; causal variants use current and past frames. EPE$_{\mathrm{3D}}$ and MTE are in meters, and Object RE is in degrees. Bold and underline mark the best and second-best values across both groups.}
\label{tab:movif-object-dynamics-reset8}
\setlength{\tabcolsep}{4pt}
\begin{tabular*}{\textwidth}{@{\extracolsep{\fill}}lrrrrrrrr@{}}
\toprule
& \multicolumn{4}{c}{Offline} & \multicolumn{4}{c}{Causal} \\
\cmidrule(lr){2-5}\cmidrule(lr){6-9}
Method & EPE$_{3D}$ $\downarrow$ & MTE $\downarrow$ & IoU $\uparrow$ & RE $\downarrow$ & EPE$_{3D}$ $\downarrow$ & MTE $\downarrow$ & IoU $\uparrow$ & RE $\downarrow$ \\
\midrule
SM4RT & \underline{0.2327} & \underline{0.1954} & \underline{0.6877} & \underline{3.96} & 0.2609 & 0.2226 & 0.6632 & 4.26 \\
ARROW & 0.3688 & 0.2839 & 0.6385 & 4.43 & 0.3091 & 0.2449 & 0.6705 & 4.55 \\
Point4D & 0.3475 & 0.2941 & 0.6115 & 4.21 & 0.3440 & 0.2883 & 0.6113 & 4.52 \\
4RC & 0.2892 & 0.2360 & 0.6712 & 4.30 & 0.2854 & 0.2321 & 0.6774 & 4.42 \\
V-DPM & 0.2904 & 0.2432 & 0.6504 & 4.24 & 0.2732 & 0.2288 & 0.6680 & 4.38 \\
L4P & 0.3854 & 0.3141 & 0.5867 & 5.55 & 0.3846 & 0.3135 & 0.5856 & 6.06 \\
TAPIP3D + DA3 & 1.2372 & 1.0291 & 0.2651 & --- & 0.9195 & 0.7074 & 0.3632 & --- \\
\midrule
MoCA4D & --- & --- & --- & --- & \textbf{0.2186} & \textbf{0.1777} & \textbf{0.7172} & \textbf{2.88} \\
\bottomrule
\end{tabular*}
\end{table*}

\section{Experiments}
\label{sec:exp}

\begin{table*}[t]
\centering\small
\caption{\textbf{3D tracking on dynamic objects.} Common scored tracks of 2,692 AV2 and 109 ADT objects, including stationary periods. All tracks start from predicted boxes without GT resets. Bold/underline rank common-input rows; Boxer additionally uses gravity. CE/CME/RME are center/center-motion/rotation-motion errors.}
\label{tab:real_world_tracking}
\label{tab:dynamic-external}
\setlength{\tabcolsep}{3pt}
\begin{tabular*}{\textwidth}{@{\extracolsep{\fill}}lrrrrrrrr@{}}
\toprule
& \multicolumn{4}{c}{AV2} & \multicolumn{4}{c}{ADT} \\
\cmidrule(lr){2-5}\cmidrule(lr){6-9}
Method & IoU $\uparrow$ & CE $\downarrow$ & CME $\downarrow$ & RME $\downarrow$ & IoU $\uparrow$ & CE $\downarrow$ & CME $\downarrow$ & RME $\downarrow$ \\
& (\%) & (m) & (m/s) & ($^\circ$/s) & (\%) & (m) & (m/s) & ($^\circ$/s) \\
\midrule
WildDet3D & 20.566 & 9.519 & 11.251 & 61.410 & 0.038 & 1.216 & 1.237 & 113.372 \\
WildDet3D + Kalman & 20.031 & 9.429 & 6.323 & 26.419 & 0.024 & 1.240 & 0.941 & \textbf{58.500} \\
OVMono3D-Lift & 8.929 & 14.110 & 9.139 & 52.184 & 1.533 & 0.820 & 0.673 & 242.610 \\
OVMono3D-Lift + Kalman & 9.443 & 13.883 & 5.577 & 21.288 & 1.499 & 0.816 & 0.457 & 112.558 \\
\midrule
MoCA4D observation & \textbf{21.559} & \underline{6.317} & 14.111 & 33.468 & 2.643 & 0.627 & 1.132 & 189.651 \\
MoCA4D observation + Kalman & \underline{20.583} & \textbf{6.228} & 7.654 & 32.191 & 2.200 & 0.621 & 0.694 & 111.070 \\
MoCA4D Star, $A$ only & 16.468 & 6.768 & \underline{4.878} & \textbf{12.402} & \underline{2.653} & \underline{0.619} & \underline{0.451} & 68.229 \\
MoCA4D Star, $A+B$ & 19.528 & 6.381 & \textbf{4.826} & \underline{12.788} & \textbf{2.784} & \textbf{0.612} & \textbf{0.449} & \underline{67.531} \\
\midrule
\multicolumn{9}{l}{Additional input: gravity direction} \\
Boxer + gravity & 0.798 & 43.660 & 6.532 & 65.140 & 7.643 & 0.320 & 0.644 & 152.410 \\
\bottomrule
\end{tabular*}
\end{table*}

\begin{table*}[t]
\begin{minipage}[t]{0.48\textwidth}
\centering\small
\caption{\textbf{Observation conditioning.} 2,000-update ablation on 10,523 validation objects. Bold marks column best.}
\label{tab:ablation-observation}
\setlength{\tabcolsep}{4pt}
\renewcommand{\arraystretch}{1.12}
\begin{tabular*}{\linewidth}{@{\extracolsep{\fill}}ccrr@{}}
\toprule
Prior map & Box attention & PAG$_{uv}$ $\downarrow$ & IoU $\uparrow$ \\
& & (px) & (\%) \\
\midrule
On  & On  & \textbf{32.11} & \textbf{14.54} \\
Off & On  & 33.45 & 14.34 \\
Off & Off & 33.15 & 14.14 \\
\bottomrule
\end{tabular*}
\end{minipage}\hfill
\begin{minipage}[t]{0.48\textwidth}
\centering\small
\caption{\textbf{Motion decomposition.} 384-update ablation; full rollouts on 147 MOVi-F scenes. Bold marks column best.}
\label{tab:ablation-motion-decomposition}
\setlength{\tabcolsep}{3pt}
\renewcommand{\arraystretch}{1.12}
\begin{tabular*}{\linewidth}{@{\extracolsep{\fill}}llrrr@{}}
\toprule
Prediction & Loss & CE $\downarrow$ & IoU $\uparrow$ & Box RE $\downarrow$ \\
& & (m) & (\%) & ($^\circ$) \\
\midrule
$A,B$ & Separate & \textbf{1.1970} & \textbf{31.72} & 26.78 \\
$A,B$ & Composed & 1.2481 & 30.89 & \textbf{26.71} \\
$AB$ & Direct & 1.2672 & 31.06 & 26.72 \\
\bottomrule
\end{tabular*}
\end{minipage}
\end{table*}

\begin{table*}[t]
\centering\small
\caption{\textbf{Star refinement controls.} 48-frame windows from 110 recordings. Left: center weights and history strength with recursion. Right: seed policies with uniform weights and $\lambda_c=0.25$. Bold marks each panel's best value.}
\label{tab:star-controls}
\begin{minipage}[t]{0.48\textwidth}
\centering
\begin{tabular*}{\linewidth}{@{\extracolsep{\fill}}lrrr@{}}
\toprule
Weights & $\lambda_c$ & CE (m) $\downarrow$ & CME (m/s) $\downarrow$ \\
\midrule
Uniform & 0.50 & 2.0814 & 1.8638 \\
Recent & 0.25 & \textbf{2.0355} & 2.4285 \\
Recent & 0.50 & 2.0654 & \textbf{1.7185} \\
\bottomrule
\end{tabular*}
\end{minipage}\hfill
\begin{minipage}[t]{0.48\textwidth}
\centering
\begin{tabular*}{\linewidth}{@{\extracolsep{\fill}}lrrr@{}}
\toprule
History seeds & CE (m) $\downarrow$ & CME (m/s) $\downarrow$ & RME ($^\circ$/s) $\downarrow$ \\
\midrule
Raw & \textbf{2.0280} & \textbf{2.5181} & 64.6550 \\
Nonrecursive & 2.0309 & 2.5470 & 56.0739 \\
Recursive & 2.0352 & 2.5406 & \textbf{41.2063} \\
\bottomrule
\end{tabular*}
\end{minipage}
\end{table*}

\subsection{Experimental Setup}
\paragraph{Baselines.}
For box lifting, we compare with DetAny3D~\cite{zhang2025detect}, OVMono3D-Lift~\cite{yao2026open}, and WildDet3D~\cite{huang2026wilddet3d}.
For object motion, we use SM4RT~\cite{lin2026sm4rt}, ARROW~\cite{fradlin2026arrow}, Point4D~\cite{jeon2026point4d}, 4RC~\cite{luo20264rc}, V-DPM~\cite{Sucar_2026_CVPR}, L4P~\cite{badki2026l4p}, and TAPIP3D~\cite{tapip3d} with Depth Anything 3~\cite{depthanything3}.
Tracking compares framewise boxes, causal Kalman filtering~\cite{kalman1960new} with rotation smoothing, and Star Optimization. Our filtering adaptation of WildDet3D uses actual elapsed time and excludes future-track corrections. We also evaluate Boxer's~\cite{boxer2026} framewise lifting with gravity. Since its tracking and fusion assume static objects in world coordinates, we use its per-frame lifting outputs to evaluate trajectories that include moving objects.

\paragraph{Datasets and inputs.}
Observation accuracy is evaluated on Omni3D~\cite{brazil2023omni3d}, rigid motion on MOVi-F~\cite{greff2022kubric}, and tracking on 150 AV2 validation recordings~\cite{wilson2021argoverse2} and 10 ADT Apartment recordings~\cite{pan2023adt}.
ADT is an external evaluation dataset used neither for pretraining nor adaptation of either MoCA4D branch.
For Omni3D, we follow the MoCA3D~\cite{jeon2026moca3d} preprocessing protocol, using tight 2D box annotations supplemented with boxes derived from SAM2~\cite{ravi2025sam2} where tight boxes are unavailable.
AV2 uses SAM 2-derived pseudo boxes, while ADT uses ground-truth 2D boxes. 
For box lifting and tracking, all methods receive the same associated 2D prompts and provided camera intrinsics.
These experiments evaluate geometry and 3D trajectories conditional on the prompts, without object detection or identity association.

\subsection{Monocular 3D Bounding Box Estimation}
We report PAG$_{uv}$~\cite{jeon2026moca3d}, the mean pixel error of projected 3D box corners after Hungarian matching at a 512-pixel reference resolution, and 3D IoU.
In Table~\ref{tab:omni3d-observation}, MoCA4D achieves the lowest PAG$_{uv}$ on ARKitScenes~\cite{baruch2021arkitscenes} and SUNRGBD~\cite{song2015sun} and the second-lowest overall on Omni3D, with competitive IoU on the two indoor datasets.
It processes 11.52 observations per second, providing $1.7\times$ the throughput of the fastest baseline with the lowest GPU memory footprint of 1.65\,GiB.

\subsection{Rigid Object Motion Estimation}
\label{sec:motion-eval}
To compare object-level SE(3) prediction from monocular video, we convert point-tracking and 4D reconstruction outputs into rigid transformations. Up to 64 shared points per object are selected from its initial ground-truth mask using an $8\times8$ box grid. Equal-weight Kabsch alignment~\cite{kabsch1976solution} fits a rigid transformation to their predicted 3D correspondences, which propagates the initial cuboid. Predicted camera poses map world-space trajectories into the current camera frame. Each baseline receives a metric scale from first-frame ground-truth depth, fixed across objects and time. TAPIP3D uses DA3-predicted depth and intrinsics; MoCA4D predicts the camera and object transformations directly.

We initialize each cuboid from ground truth and reset it every eight frame transitions (GT reset-8), measuring motion accuracy within fixed windows while limiting accumulated drift. Causal methods use current and past frames; offline methods can access future frames.
EPE$_{\mathrm{3D}}$~\cite{gu2019hplflownet} averages Euclidean errors over the eight corresponding cuboid corners, and MTE takes the temporal median for each corner before averaging. We also report 3D IoU and Object RE, the geodesic error of inter-frame object rotation.

MoCA4D achieves the best values across all four metrics among the causal and offline methods in Table~\ref{tab:movif-object-dynamics-reset8}. Relative to offline SM4RT, IoU rises from 0.6877 to 0.7172 and Object RE falls from $3.96^\circ$ to $2.88^\circ$. In full rollouts without resets, MoCA4D retains higher IoU (0.5683 versus 0.5621) and lower Object RE ($2.95^\circ$ versus $4.20^\circ$), while SM4RT has lower corner errors. Table~\ref{tab:movif-object-dynamics} gives the full-rollout results for all methods.

\paragraph{Object motion on real data.}
For each frame pair, we propagate the previous ground-truth cuboid to the next frame using the same predicted $A$ and either predicted $B$ or $B=I$.
On dynamic frame pairs from AV2 and ADT, predicted $B$ reduces displacement and corner propagation errors, although AV2 rotation error increases.
Predicted $A$ reduces relative camera translation and rotation errors compared with $A=I$.
Appendix~\ref{sec:supp-real-motion} gives the one-step evaluation protocol, uncertainty estimates, and results on HOT3D and HOI4D.

\subsection{3D Bounding Box Tracking}
\label{sec:tracking-eval}
The complete pipeline starts from predicted boxes and updates them causally without ground-truth resets. Fig.~\ref{fig:qual} presents qualitative comparisons in which Star Optimization produces more consistent box orientations across frames than framewise predictions, for both static and moving objects under camera motion. Appendix~\ref{sec:supp-qual} shows additional AV2 and HOT3D sequences.
We measure 3D IoU, Euclidean center error (CE), center motion error (CME), and rotation motion error (RME). CME measures changes in world-coordinate center error per second; RME measures disagreement with ground-truth inter-frame cuboid rotation per second after equivalent axis alignment. CE and IoU complement these motion metrics by capturing absolute localization error, including constant position bias.

\paragraph{Dynamic-object comparison.}
Table~\ref{tab:real_world_tracking} compares external baselines and matched controls on 2,692 AV2 and 109 ADT dynamic objects. We select objects with GT speed $\geq0.05$\,m/s or reliable rotation speed $\geq5^\circ$/s sustained for 0.5\,s.
We evaluate all eligible observations, including stationary periods. Appendix~\ref{sec:supp-dynamic-tracking} extends this comparison to HOT3D and HOI4D.

On AV2, Star Optimization has lower CE, CME, and RME than both external Kalman baselines, but lower IoU than WildDet3D + Kalman. On ADT, it achieves the highest IoU and lowest CE/CME among common-input methods; WildDet3D + Kalman has lower RME.
Boxer with additional gravity has higher ADT IoU and lower CE.

\paragraph{Camera and object motion in tracking.}
Table~\ref{tab:real_world_tracking} compares Star Optimization using $A+B$ with camera-only refinement ($B=I$) and Kalman filtering of the same MoCA4D observations.
Checkpoints and observations remain fixed, and each variant uses its own predicted states without GT.
Relative to camera-only refinement, adding $B$ increases IoU by 3.060 percentage points on AV2 and 0.131 on ADT and lowers CE and CME on both datasets, although AV2 RME increases.
Compared with Kalman filtering, Star Optimization lowers CME and RME on both datasets.
AV2 has lower IoU and higher CE, while ADT improves across all four metrics.

\paragraph{All-object tracking and localization.}
Across all evaluated objects, including static objects, Star Optimization reduces CME by 38.1\% on AV2 and 38.7\% on ADT and lowers RME relative to Kalman filtering of the same MoCA4D observations.
A separate diagnostic on ADT shows that replacing only the predicted center with GT yields a much larger IoU increase than replacing size or rotation, indicating that center localization remains a major limitation.
Table~\ref{tab:all-object-tracking} reports the all-object comparison, and Table~\ref{tab:adt-oracle} breaks down the ADT result by box component.

\subsection{Ablation Studies}
\label{sec:ablations}
\paragraph{Observation conditioning.}
We train all variants for 2,000 updates on the same 60,000-object subset, with matched initialization, frozen backbone, and seed. These ablation checkpoints are separate from the benchmark models in Table~\ref{tab:omni3d-observation}; scores are macro-averaged across the Omni3D datasets on 10,523 validation objects.
Full conditioning gives the lowest projected-corner error and highest IoU (Table~\ref{tab:ablation-observation}). Removing the prior map raises PAG$_{uv}$ by 1.34 pixels; removing both the prior and box attention raises it by 1.04 pixels and lowers IoU by 0.40 percentage points.

\paragraph{Motion decomposition and supervision.}
We compare three variants: predicting $A$ and $B$ with separate supervision; predicting $A$ and $B$ with direct motion supervision applied only to their composition $AB$; and directly predicting and supervising $AB$ without separate $A$ and $B$ outputs.
All variants share the same architecture and training settings except for the motion output heads and their supervision.
We train each variant for 384 updates on the same 1,024 MOVi-F scenes and evaluate  full rollouts on 147 validation scenes.
In Table~\ref{tab:ablation-motion-decomposition}, separate supervision yields the lowest CE and highest IoU while its box RE is slightly higher than that of the two alternatives.

\paragraph{Star refinement.}
Table~\ref{tab:star-controls} isolates weighting and feedback. At $\lambda_c=0.5$, recent weights lower CE and CME; increasing $\lambda_c$ trades higher CE for lower CME. Raw seeds propagate uncorrected boxes; nonrecursive seeds apply one correction; recursive seeds reuse refined states. With uniform weights and $\lambda_c=0.25$, recursion lowers RME relative to both alternatives. Appendix~\ref{sec:supp-star} compares history lengths, fusion weights, and feedback policies.

\section{Conclusion}
We introduce MoCA4D for monocular 3D bounding box lifting and tracking with explicit camera and object motion. 
Star Optimization connects these predictions over time by fusing current observations with propagated history, enabling causal tracking of moving objects.
Experiments demonstrate efficient box lifting and more accurate center localization than the external Kalman baselines on dynamic-object tracks in AV2 and ADT.
MoCA4D also yields lower center motion errors on both datasets and lower rotation motion errors on AV2.

MoCA4D relies on associated 2D box prompts, leaving object detection outside its scope.
Its rigid motion formulation captures each object's overall movement but does not model internal articulation or deformation.

\section*{LLM Usage Statement}
\addcontentsline{toc}{section}{LLM Usage Statement}
We used OpenAI Codex, powered by OpenAI Astra, to assist with assessing research ideas, planning experiments, and implementing training and experimental code. It also supported manuscript preparation through sentence-level revisions and writing assistance. The authors take full responsibility for all scientific claims, experimental results, and the final manuscript.

\begingroup
\fontsize{9}{10}\selectfont
\setlength{\bibsep}{0.4pt plus 0.4pt}
\renewcommand{\bibsection}{\section*{\refname}\addcontentsline{toc}{section}{\refname}}
\bibliographystyle{ieeenat_fullname}
\bibliography{main}
\endgroup

\clearpage
\appendix
\setcounter{table}{0}
\setcounter{figure}{0}
\setcounter{equation}{0}
\renewcommand{\thetable}{S\arabic{table}}
\renewcommand{\thefigure}{S\arabic{figure}}
\renewcommand{\theequation}{S\arabic{equation}}
\renewcommand{\theHsection}{supp.\Alph{section}}
\renewcommand{\theHtable}{supp.\arabic{table}}
\renewcommand{\theHfigure}{supp.\arabic{figure}}
\renewcommand{\theHequation}{supp.\arabic{equation}}
\setcounter{dbltopnumber}{3}
\setcounter{topnumber}{3}
\setcounter{totalnumber}{5}
\setlength{\dblfloatsep}{10pt plus 2pt minus 2pt}
\setlength{\dbltextfloatsep}{12pt plus 2pt minus 2pt}

\twocolumn[{
  \begin{minipage}{\textwidth}
    \pdfbookmark[0]{Supplementary Material}{supplement}
    \centering
    {\Large\bfseries Supplementary Material\par}
    \medskip
    \includegraphics[width=\textwidth]{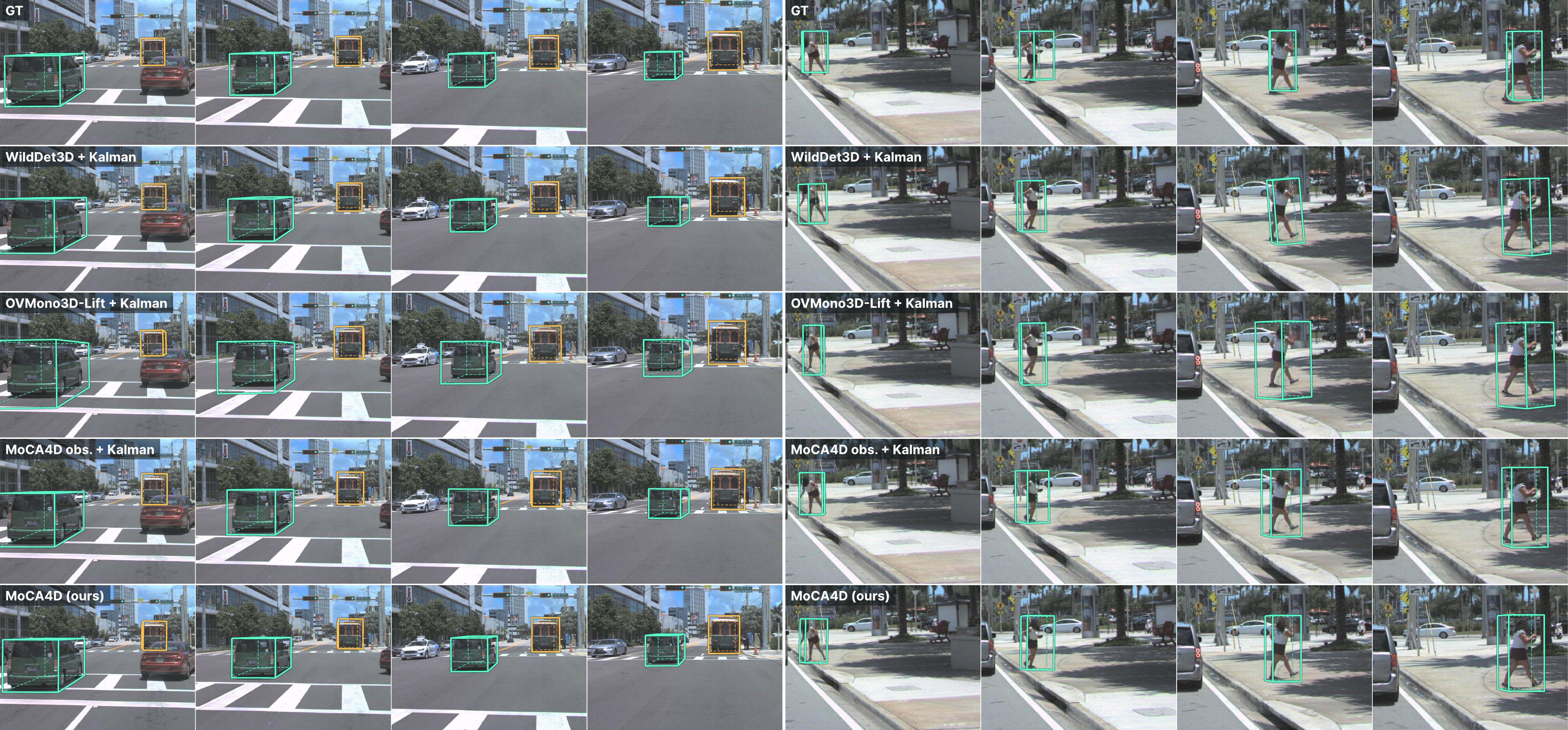}
    \captionsetup{hypcap=false}
    \captionof{figure}{\textbf{Qualitative comparisons with baselines.} The left sequence follows a minivan (green) and a bus (orange) at an intersection. The right sequence follows a pedestrian on the sidewalk. Within each sequence, columns advance in time and the top row shows ground truth.}
    \label{fig:supp-qual-av2}
  \end{minipage}
  \medskip
}]

\noindent Appendices~\ref{sec:supp-implementation}--\ref{sec:supp-eval-protocols} describe training, illustrate tracking on AV2 and HOT3D, and specify the evaluation protocols. Appendices~\ref{sec:supp-rollout} and~\ref{sec:supp-real-motion} extend the motion experiments in Sec.~\ref{sec:motion-eval} to full MOVi-F rollouts and one-step predictions on real videos. Appendices~\ref{sec:supp-all-tracking}--\ref{sec:supp-adt-diagnosis} expand the tracking results in Sec.~\ref{sec:tracking-eval} and isolate sources of localization error on ADT. Appendix~\ref{sec:supp-star} examines the role of history, weighting, and feedback in the Star ablations of Sec.~\ref{sec:ablations}.

\begin{figure*}[!t]
\centering
\includegraphics[width=\textwidth]{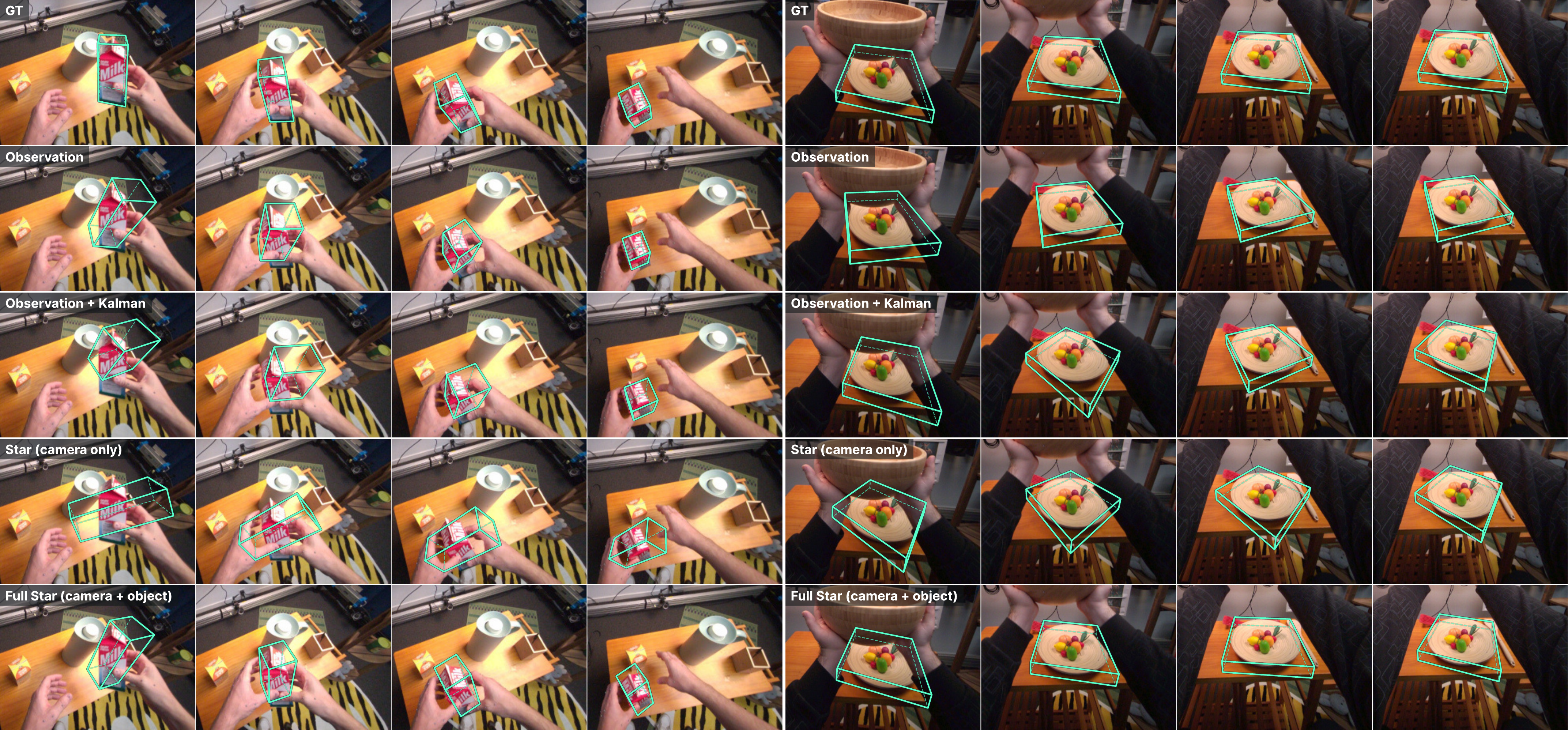}
\caption{\textbf{Qualitative comparison of motion modeling in 3D box tracking.}
On HOT3D, we compare raw and Kalman-filtered box observations with two Star variants.
One uses camera motion alone ($A$), and the other uses both camera and object motion ($A+B$).
A hand tilts a milk carton in the left sequence. The tracked object on the right is the bamboo plate beneath the bowl.
The top row shows the ground-truth boxes for each sequence.}
\label{fig:supp-qual-hot3d}
\end{figure*}

\section{Implementation and training details}
\label{sec:supp-implementation}
Both branches share a frozen DA3-Large-1.1 image encoder~\cite{depthanything3} with 24 transformer blocks.
The encoder takes a $518\times518$ RGB frame as input.
It divides the frame into non-overlapping $14\times14$-pixel patches, forming a $37\times37$ grid.
For the current frame, the observation branch reads 1,024-dimensional patch features from blocks 12, 16, 20, and 24.
The dynamic branch reads the block~12 patch features from each of two consecutive frames.
The shared encoder accounts for 304.37M of the 396.35M model parameters.
We train the observation branch (56.78M) and the dynamic branch (35.20M) separately.

\subsection{Observation architecture and supervision}
\label{sec:supp-observation}
The observation branch reads 1,024-channel features from encoder blocks 12, 16, 20 and 24. We adapt DA3's multi-frame design to MoCA4D by extracting these features from each frame independently. A box-conditioned DPT fuses these features into a $129\!\times\!129$ spatial grid with 256 channels. The prompt encoder has three layers of width 256. For each prompted object, one query passes through two decoder layers, each with eight attention heads and a feed-forward width of 2,048. Each layer has its own geometry and confidence heads.

We supervise the encoded center offset, depth, dimensions,
and rotation with $\ell_1$ losses, and train the confidence
head using binary cross-entropy with soft quality targets
based on depth accuracy and box overlap.
Supervision is applied to every decoder layer:
\begin{equation}
\begin{aligned}
\mathcal{L}_{\mathrm{obs}}
=
\sum_{\ell=1}^{L}
\bigl(
&\mathcal{L}_{uv}^{(\ell)}
+\mathcal{L}_{z}^{(\ell)}
+\mathcal{L}_{s}^{(\ell)} \\
&+\mathcal{L}_{R}^{(\ell)}
+0.5\,\mathcal{L}_{\mathrm{conf}}^{(\ell)}
\bigr),
\end{aligned}
\end{equation}

\noindent where $L=2$.
During adaptation, rotation and dimension targets are
jointly selected from the 24 equivalent cuboid
parameterizations to minimize their combined regression
loss. At inference, we use the final decoder output. The confidence head is trained with the observation branch; its output does not determine the fixed Star fusion weights.

\subsection{Correspondence features and motion decoder}
Given patch features $\mathbf{F}_{t-1}$ and $\mathbf{F}_t$, a cross-frame transformer alternates within-frame self-attention and bidirectional cross-attention.
The matcher has two blocks of width 128, each with four attention heads and a feed-forward width of 512. Following the global matching formulation of GMFlow~\cite{xu2022gmflow}, the resulting features $\mathbf{H}_{t-1}$ and $\mathbf{H}_t$ define soft correspondences:
\begin{equation}
\begin{aligned}
&\mathbf{P}_t
= \operatorname{softmax}_{\mathrm{row}}\!\left(
\mathbf{H}_{t-1}\mathbf{H}_t^\top / \sqrt{d}
\right), \\[4pt]
&\hat{\mathbf{v}}_t(a)
= \sum_b P_t(a,b)
\left(\mathbf{g}_b-\mathbf{g}_a\right).
\end{aligned}
\label{eq:dynamic-matching}
\end{equation}

\noindent where $d$ is the feature dimension and $\mathbf{g}_a$ denotes a normalized patch coordinate.
The expected displacement $\hat{\mathbf{v}}_t(a)$ provides a differentiable motion cue.
For each patch, we construct a descriptor combining appearance features with cross-frame matching information. Descriptors are computed in both temporal directions and projected back into the backbone feature space:
\begin{equation}
\tilde{\mathbf{F}}_s
=
\mathbf{F}_s+\phi(\mathbf{D}_s),
\qquad s\in\{t-1,t\},
\label{eq:dynamic-feature}
\end{equation}

\noindent where $\phi$ is an MLP.
This residual update incorporates correspondence information into the features used for motion prediction.
We construct spatial tokens by applying $7\times7$ RoI Align to the enhanced features within each object's 2D boxes in both frames.
The pooled features are projected and combined with geometric and temporal embeddings, yielding 98 tokens $\mathbf{Z}_t^i$. The motion decoder has six pre-normalized blocks of width 512, with eight attention heads and a feed-forward width of 2,048. It uses four register tokens, eight camera tokens and eight object tokens. In every block, the object tokens also attend to the 98 pooled tokens for that object.

\subsection{Motion parameterization and targets}
We assemble these predictions into the camera and object
transformations:
\begin{equation}
\begin{aligned}
&\hat{\mathbf{A}}_t =
\begin{bmatrix}
\hat{\mathbf{R}}_t^A & \hat{\mathbf{t}}_t^A \\
\mathbf{0}^{\top} & 1
\end{bmatrix}, \\[4pt]
&\hat{\mathbf{B}}_t^i =
\begin{bmatrix}
\hat{\mathbf{R}}_t^{B,i} & \hat{\mathbf{t}}_t^{B,i} \\
\mathbf{0}^{\top} & 1
\end{bmatrix}.
\end{aligned}
\label{eq:predicted-motion}
\end{equation}

\noindent Their composition
$\hat{\mathbf{A}}_t\hat{\mathbf{B}}_t^i$
propagates the object's geometry into the current camera
frame according to Eq.~\eqref{eq:motion-propagation}.

The displacement target is evaluated at the same supplied center:
\begin{equation}
\mathbf{d}_t^{i,*}
=
(\mathbf{R}_t^{B,i,*}-\mathbf{I})
\bar{\mathbf{c}}_{t-1}^i+\mathbf{t}_t^{B,i,*}.
\end{equation}

\subsection{Training settings}
The observation branch is pretrained on Omni3D~\cite{brazil2023omni3d}, which contains KITTI~\cite{Geiger2012CVPR}, nuScenes~\cite{nuscenes2019}, Objectron~\cite{ahmadyan2021objectron}, SUN RGBD~\cite{song2015sun}, ARKitScenes~\cite{baruch2021arkitscenes}, and Hypersim~\cite{roberts2021hypersim}. The dynamic branch is pretrained on MOVi-F~\cite{greff2022kubric} sequences.
We then adapt both branches on HOT3D~\cite{banerjee2025hot3d}, HOI4D~\cite{liu2022hoi4d}, Argoverse 2~\cite{wilson2021argoverse2} and Virtual KITTI 2~\cite{cabon2020virtual}.

We use AdamW throughout training, with initial pretraining learning rates of $10^{-4}$ for the observation branch and $5\times10^{-5}$ for the dynamic branch.
During adaptation, the base learning rates are $2\times10^{-5}$ and $3\times10^{-5}$ for the observation
and dynamic branches, respectively, with $5\times10^{-5}$ for their prediction heads. For updates 4,501 through 5,000 we reduce both dynamic learning rates tenfold and retain the optimizer moments. The rollout loss has weight 0.1. During adaptation we apply the flow loss (weight 1) on MOVi-F and Virtual KITTI 2.

\par\addvspace{10pt}
\noindent\begin{minipage}{\columnwidth}
\small
\textbf{Training budget for the reported models.}\par
\vspace{4pt}
\renewcommand{\arraystretch}{1.10}
\begin{tabular*}{\linewidth}{@{\extracolsep{\fill}}lrrr@{}}
\toprule
 & & \multicolumn{2}{c}{Per epoch} \\
\cmidrule(l){3-4}
Stage and branch & Epochs & Updates & Examples \\
\midrule
Pretrain observation & 8 & 7,414 & 1,067,612 objects \\
Pretrain dynamics & 25 & 180 & 5,737 scenes \\
\addlinespace[2pt]
Adapt observation & 5 & 3,707 & 533,808 objects \\
Adapt dynamics & 10 & 500 & 16,000 clips \\
\bottomrule
\end{tabular*}
\end{minipage}\par\addvspace{10pt}

A pretraining epoch covers the eligible Omni3D annotations or MOVi-F training scenes once. The final batches contain 140 objects and nine scenes, respectively. Adaptation uses a fixed number of optimizer updates per epoch. Its example counts include repeated draws from the training mixture. Tracking uses the EMA weights at the end of each adaptation schedule. During adaptation, the sampling proportions for HOT3D, HOI4D, AV2, and Virtual KITTI 2 are 20, 12, 20, and 8 percent. The remaining 40 percent comes from Omni3D for observation training and MOVi-F for dynamic training. Full batches contain 144 objects for observation or 32 scenes for dynamic pretraining. Each dynamic adaptation update uses 32 clips of 24 frames, with up to four objects per clip. The observation and dynamic branches use weight decay of $10^{-4}$ and $0.01$, gradient norm limits of $0.1$ and $1.0$, and EMA decay of $0.999$ and $0.99$, respectively.

\section{Qualitative tracking results}\label{sec:supp-qual}
Figure~\ref{fig:supp-qual-av2} compares full MoCA4D with Kalman-filtered boxes from WildDet3D, OVMono3D-Lift and our observation branch. In the last two frames at right, MoCA4D follows the pedestrian's annotated box width more closely than OVMono3D-Lift + Kalman.

Figure~\ref{fig:supp-qual-hot3d} compares raw and Kalman-filtered observations with camera-only Star ($A$) and full Star ($A+B$). In the second frame of each sequence, adding $B$ brings the carton's long axis and the edges of the plate box closer to their annotated orientations.

\section{Evaluation data and protocols}\label{sec:supp-eval-protocols}

\subsection{Datasets and splits}
ADT is an external evaluation dataset used neither for pretraining nor adaptation of either MoCA4D branch. AV2 is included in branch adaptation and is evaluated on its validation recordings. Table~\ref{tab:data-roles} summarizes the training roles of the principal evaluation datasets; the complete training configuration is given in Sec.~\ref{sec:supp-implementation}. AV2 prompts are SAM~2~\cite{ravi2025sam2}-derived pseudo boxes, while ADT prompts are dataset ground-truth 2D boxes.

\begin{table*}[!tbp]
\centering\small
\caption{Training roles and evaluation sets. ADT is excluded from MoCA4D branch pretraining and adaptation.}
\label{tab:data-roles}
\begin{tabular*}{\textwidth}{@{\extracolsep{\fill}}lp{.37\textwidth}p{.43\textwidth}@{}}
\toprule
Data & MoCA4D branch training & Evaluation \\
\midrule
Omni3D & Observation pretraining & Conditioning ablation on 10,523 validation objects across six datasets \\
MOVi-F & Dynamic pretraining & 147 validation scenes (1,750 objects) \\
AV2 & Branch adaptation & 150 validation recordings \\
ADT & Neither pretraining nor adaptation & 10 Apartment recordings \\
\bottomrule
\end{tabular*}
\end{table*}

For the conditioning ablation (Table~\ref{tab:ablation-observation}), all variants share the initialization, frozen backbone, seed, 60,000-object training subset and 2,000 updates. We average within each of the six validation datasets and then across datasets (10,523 objects total).

\subsection{Inputs for motion estimation and tracking}

\paragraph{MOVi-F motion baselines.}
For MOVi-F, point and reconstruction baselines receive a first-frame GT mask for point selection and a first-frame GT depth scale held fixed across objects and resets. Offline variants access future frames; causal variants access current and past frames. TAPIP3D uses DA3-predicted depth and intrinsics. These adaptations differ from the prompt-conditioned trajectory task.

\paragraph{Tracking protocol.}
We evaluate 150 AV2 validation recordings and 10 ADT Apartment recordings on common prepared object--time support. The final scored support contains 69,434 box events and 60,119 motion pairs on AV2, and 431,681 box events and 401,980 motion pairs on ADT. Observations are acquired every two raw frames; the dynamic branch also consumes intermediate frames. Scoring starts after four observations are available. Tracks are initialized from predicted boxes and receive no ground-truth resets. We average within each object, then across objects within a recording, and finally across recordings within each dataset. When Star cannot compute a valid refinement, it returns the current 3D box predicted by the observation branch. We count this as one fallback and include the box in the tracking metrics.

\paragraph{WildDet3D-style filtering.}
For the observation-only Kalman rows in Table~\ref{tab:real_world_tracking}, we apply the same causal filter to boxes from WildDet3D, OVMono3D-Lift and MoCA4D. We adapt the WildDet3D demo~\cite{huang2026wilddet3d} to the measured time between observations.

The state comprises the camera-frame center $c$, box dimensions $s$ and center velocity $v$. Between observations, the filter advances the center by $\Delta t\,v$ and carries $s$ and $v$ forward. Every two raw frames it updates the state from the observed center and dimensions. The first box sets $c$ and $s$, with $v=0$.

For $I=I_3$, the initial, process and measurement covariances are
\[
\begin{aligned}
 P_0 &= \operatorname{diag}(I,I,10I),\\
 Q &= \operatorname{diag}(0.25I,0.01I,I),\\
 R &= \operatorname{diag}(I,0.25I).
\end{aligned}
\]
$Q$ and $R$ are independent of $\Delta t$. For rotation, we average the previous filtered and current observed unit quaternions with equal weights and normalize the result, retaining their original signs.

\paragraph{Gravity input.}
Boxer~\cite{boxer2026} uses its official framewise lifting configuration, with gravity extracted from dataset camera poses as an additional input. It receives neither camera translation nor world heading, depth, semi-dense points, or object 3D ground truth. We do not enable Boxer's optional tracking or fusion, and report it in a separate input group.

\section{Full-rollout object motion}\label{sec:supp-rollout}
\begin{table*}[!tbp]
    \centering
    \small
    \renewcommand{\arraystretch}{1.08}
    \caption{\textbf{Rigid object motion on MOVi-F with full rollout.}
    Full rollouts are evaluated on 1,750 objects from 147 validation scenes.
    Bold and underlined values indicate the best and second-best results,
    respectively, across all rows with reported values.
    }
    \label{tab:movif-object-dynamics}

    \begin{tabular*}{\textwidth}{@{\extracolsep{\fill}}lrrrr@{}}
        \toprule
        Method
         & EPE$_{\mathrm{3D}}$ (m) $\downarrow$
         & MTE (m) $\downarrow$
         & IoU $\uparrow$
         & Object RE (${}^\circ$) $\downarrow$\\
        \midrule

        \multicolumn{5}{@{}l}{
            \textbf{Offline baselines (future frames available)}
        } \\

        SM4RT~\cite{lin2026sm4rt}
         & \textbf{0.4386}  & \textbf{0.4579}
         & \underline{0.5621}  & \underline{4.20}\\

        ARROW~\cite{fradlin2026arrow}
         & 0.7295  & 0.7530  & 0.4548  & 4.65\\

        Point4D~\cite{jeon2026point4d}
         & 0.7943  & 0.8560  & 0.4270  & 4.40\\

        4RC~\cite{luo20264rc}
         & 0.6251  & 0.6666  & 0.5164  & 4.66\\

        V-DPM~\cite{Sucar_2026_CVPR}
         & 0.5906  & 0.6219  & 0.4902  & 4.62\\

        L4P~\cite{badki2026l4p}
         & 0.7887  & 0.8189  & 0.4167  & 6.16\\

        TAPIP3D~\cite{tapip3d} + DA3~\cite{depthanything3}
         & 2.1853  & 2.2053  & 0.1472  & ---\\

        \midrule
        \multicolumn{5}{@{}l}{
            \textbf{Causal baselines (current and past frames)}
        } \\

        SM4RT
         & 0.4650  & \underline{0.4710}
         & 0.5407  & 4.67\\

        ARROW
         & 0.6489  & 0.6615  & 0.4975  & 5.43\\

        Point4D
         & 0.7743  & 0.8167  & 0.4375  & 4.96\\

        4RC
         & 0.6115  & 0.6409  & 0.5281  & 4.87\\

        V-DPM
         & 0.5577  & 0.5727  & 0.5118  & 4.89\\

        L4P
         & 0.7801  & 0.7761  & 0.4195  & 7.57\\

        TAPIP3D + DA3
         & 1.8963  & 1.9300  & 0.2357  & ---\\

        \midrule
        \multicolumn{5}{@{}l}{\textbf{Ours (causal)}} \\

        MoCA4D
         & \underline{0.4594}  & 0.4820
         & \textbf{0.5683}  & \textbf{2.95}\\

        \bottomrule
    \end{tabular*}
\end{table*}

\paragraph{Long-horizon propagation.}
MoCA4D propagates each cuboid by composing the predicted camera and object motions using only current and past frames. With a ground-truth reset every eight transitions, it leads on all four MOVi-F metrics (Table~\ref{tab:movif-object-dynamics-reset8}). When these updates carry the initial cuboid through the full rollout without resets, MoCA4D retains the highest 3D IoU (0.5683) and the lowest inter-frame object rotation error ($2.95^\circ$) among the methods in Table~\ref{tab:movif-object-dynamics}. Offline SM4RT yields lower corner errors on full rollouts.

\par

\section{Real-data one-step motion diagnostics}
\label{sec:supp-real-motion}
\paragraph{Question and protocol.}
On adjacent frames from real videos, we compare predicted object motion $B$ with a stationary-object baseline ($B=I$), holding the previous GT cuboid and predicted camera motion $A$ fixed for corner propagation (Table~\ref{tab:real-b-errors}). The object head receives the previous GT center. Absolute one-step errors cannot be directly compared with MOVi-F GT reset-8 errors. HOT3D and HOI4D contribute validation sequences from domains used for branch adaptation.

\paragraph{Motion subsets.}
A pair is dynamic when GT world translation speed is at least 0.05\,m/s or reliable rotation speed is at least $5^\circ$/s. Static pairs have translation speed at most 0.025\,m/s and rotation speed at most $2.5^\circ$/s; the intervening band is ambiguous. An annotation-symmetry flag prevents rotation alone from labeling an object dynamic. Rotation errors below are raw SO(3) geodesic errors on the selected pairs, so translation-selected symmetric objects can still contribute. Thresholds are operational definitions, not native dataset labels. GT-only calibration on ADT annotated-static pairs gave a 99th-percentile translation speed of zero and rotation speed of $3.14\times10^{-5}$ degrees/s. This calibration does not validate annotation noise in every dataset. The dynamic-pair diagnostic here has no sustained-duration requirement and is distinct from the whole-track dynamic-object selection in Sec.~\ref{sec:supp-dynamic-tracking}.

\begin{table*}[!tbp]
\centering\small
\caption{\textbf{Direct object motion and controlled propagation.} Each cell is identity $B\rightarrow$ predicted $B$; lower is better. Both arms use the previous GT center, and propagation uses the previous GT cuboid with a common predicted $A$. Values are object--recording--dataset macro averages.}
\label{tab:real-b-errors}
\setlength{\tabcolsep}{4pt}
\begin{tabular*}{\textwidth}{@{\extracolsep{\fill}}lrrr@{}}
\toprule
Dataset & Displacement (mm) $\downarrow$ & Corner propagation (mm) $\downarrow$ & Rotation ($^\circ$) $\downarrow$ \\
\midrule
\multicolumn{4}{l}{\textit{Dynamic frame pairs}} \\
AV2 & $307.55\rightarrow 244.09$ & $365.33\rightarrow 298.62$ & $0.217\rightarrow 0.305$ \\
ADT & $24.83\rightarrow 22.41$ & $21.55\rightarrow 17.57$ & $2.457\rightarrow 2.191$ \\
HOT3D & $16.78\rightarrow 12.66$ & $19.34\rightarrow 14.91$ & $5.302\rightarrow 4.458$ \\
HOI4D & $7.10\rightarrow 4.92$ & $6.62\rightarrow 4.35$ & $0.872\rightarrow 0.807$ \\
\midrule
\multicolumn{4}{l}{\textit{Static frame pairs}} \\
AV2 & $1.59\rightarrow 31.67$ & $85.70\rightarrow 104.97$ & $0.069\rightarrow 0.144$ \\
ADT & $0.00\rightarrow 3.22$ & $30.44\rightarrow 31.21$ & $0.003\rightarrow 0.191$ \\
HOT3D & $0.07\rightarrow 1.64$ & $4.46\rightarrow 4.62$ & $0.045\rightarrow 0.190$ \\
HOI4D & $1.14\rightarrow 1.68$ & $1.62\rightarrow 1.79$ & $0.115\rightarrow 0.152$ \\
\bottomrule
\end{tabular*}
\end{table*}

\begin{table*}[!tbp]
\centering\small
\caption{\textbf{Dynamic-pair differences: predicted $B$ minus identity.} Negative values favor predicted motion. Brackets are individual paired 95\% recording-bootstrap intervals (2,000 resamples, seed 20261115); no multiplicity correction is applied.}
\label{tab:real-b-ci}
\setlength{\tabcolsep}{4pt}
\begin{tabular*}{\textwidth}{@{\extracolsep{\fill}}lrrr@{}}
\toprule
Dataset & $\Delta$Displacement (mm) & $\Delta$Corner (mm) & $\Delta$Rotation ($^\circ$) \\
\midrule
AV2 & -63.46 [-78.68, -49.17] & -66.72 [-81.68, -52.00] & +0.088 [+0.081, +0.096] \\
ADT & -2.41 [-3.04, -1.73] & -3.98 [-4.45, -3.49] & -0.266 [-0.351, -0.179] \\
HOT3D & -4.11 [-4.91, -3.38] & -4.43 [-5.06, -3.83] & -0.844 [-1.014, -0.664] \\
HOI4D & -2.18 [-2.33, -2.04] & -2.27 [-2.42, -2.12] & -0.065 [-0.077, -0.054] \\
\bottomrule
\end{tabular*}
\end{table*}

The dynamic one-step samples comprise 144,247 AV2, 10,771 ADT, 26,807 HOT3D, and 95,784 HOI4D object--frame pairs; static samples comprise 31,430, 999,093, 114,000, and 6,382 pairs, respectively. They come from 150, 10, 30, and 443 original recordings. Camera evaluation uses 22,543, 14,128, 30,486, and 131,122 frame pairs, respectively.

\paragraph{Effect of predicted object motion.}
On dynamic frame pairs, using the predicted object transform $B$ in place of identity reduces both displacement error and next-frame corner error on all four datasets (Table~\ref{tab:real-b-errors}). On AV2, corner error falls from 365 to 299\,mm even though rotation error rises by $0.088^\circ$. Rotation error falls on the other three datasets (Table~\ref{tab:real-b-ci}). On static pairs, the identity transform gives lower errors on all three measures across all four datasets. For moving objects, $B$ captures part of the displacement that camera motion alone leaves unexplained.

\paragraph{Frame-to-frame camera motion.}
For each adjacent frame pair, we compare the predicted camera transform $A$ with the ground-truth relative pose $C_t^{-1}C_{t-1}$. We measure the Euclidean distance between the translation vectors and the angle between the rotations. Each frame pair counts once. We average errors within each recording and give each recording equal weight within its dataset. For a stationary-camera prediction ($A=I$), the two errors equal the magnitudes of the camera's translation and rotation between frames. MoCA4D reduces both errors on all four datasets (Table~\ref{tab:real-camera}). Translation error falls from 559 to 60\,mm on AV2, and rotation error from $2.13^\circ$ to $0.53^\circ$ on ADT.

\begin{table}[!htb]
\centering\small
\caption{\textbf{One-step relative camera error.} Each cell gives the error for the stationary-camera baseline followed by the error for MoCA4D's predicted $A$.}
\label{tab:real-camera}
\setlength{\tabcolsep}{4pt}
\begin{tabular*}{\columnwidth}{@{\extracolsep{\fill}}lrr@{}}
\toprule
Dataset & Translation (mm) $\downarrow$ & Rotation ($^\circ$) $\downarrow$ \\
\midrule
AV2 & $558.841\rightarrow 59.976$ & $0.262\rightarrow 0.088$ \\
ADT & $33.203\rightarrow 22.525$ & $2.127\rightarrow 0.530$ \\
HOT3D & $7.460\rightarrow 4.789$ & $1.263\rightarrow 0.234$ \\
HOI4D & $6.074\rightarrow 4.643$ & $0.412\rightarrow 0.254$ \\
\bottomrule
\end{tabular*}
\end{table}

\section{Tracking on all objects}\label{sec:supp-all-tracking}
Table~\ref{tab:all-object-tracking} extends the dynamic-object evaluation in Table~\ref{tab:real_world_tracking} to all scored tracks on AV2 and ADT, including objects that remain stationary. At each scored frame, 3D IoU is the intersection volume of the predicted and ground-truth cuboids divided by their union volume. Center error (CE) is the Euclidean distance between their centers. Center-motion error (CME) and rotation-motion error (RME) measure errors in displacement and rotation between consecutive scored observations, per second of elapsed time.

\begin{table*}[!tbp]
\centering\small
\caption{\textbf{Tracking on all scored AV2 and ADT objects.} Tracks start from predicted boxes and receive no ground-truth resets. We score all methods on the same objects at the same times. Bold and underlined values mark the best and second-best results among methods with the same inputs. Boxer also uses gravity.}
\label{tab:all-object-tracking}
\setlength{\tabcolsep}{3pt}
\begin{tabular*}{\textwidth}{@{\extracolsep{\fill}}lrrrrrrrr@{}}
\toprule
& \multicolumn{4}{c}{AV2} & \multicolumn{4}{c}{ADT} \\
\cmidrule(lr){2-5}\cmidrule(lr){6-9}
Method & IoU $\uparrow$ & CE $\downarrow$ & CME $\downarrow$ & RME $\downarrow$ & IoU $\uparrow$ & CE $\downarrow$ & CME $\downarrow$ & RME $\downarrow$ \\
& (\%) & (m) & (m/s) & ($^\circ$/s) & (\%) & (m) & (m/s) & ($^\circ$/s) \\
\midrule
WildDet3D & 16.591 & 10.892 & 11.683 & 67.261 & 0.312 & 1.519 & 1.942 & 153.160 \\
WildDet3D + Kalman & 16.074 & 10.774 & 6.632 & 29.662 & 0.168 & 1.580 & 1.516 & \textbf{65.108} \\
OVMono3D-Lift & 7.204 & 15.154 & 9.673 & 62.706 & 0.239 & 1.943 & 1.200 & 249.445 \\
OVMono3D-Lift + Kalman & 7.667 & 14.876 & \underline{5.902} & \underline{24.161} & 0.240 & 1.931 & \textbf{0.759} & 106.620 \\
\midrule
MoCA4D observation & \textbf{17.826} & 6.909 & 14.861 & 36.139 & \textbf{0.932} & 1.274 & 2.759 & 202.176 \\
MoCA4D observation + Kalman & \underline{16.867} & \textbf{6.822} & 8.179 & 39.493 & 0.587 & \underline{1.252} & 1.640 & 112.628 \\
MoCA4D Star Optimization & 16.610 & \underline{6.893} & \textbf{5.059} & \textbf{14.480} & \underline{0.915} & \textbf{1.212} & \underline{1.004} & \underline{74.363} \\
\midrule
\multicolumn{9}{l}{Additional input: gravity direction} \\
Boxer + gravity & 0.590 & 44.481 & 6.430 & 65.430 & 6.173 & 0.469 & 1.211 & 128.962 \\
\bottomrule
\end{tabular*}
\end{table*}

On AV2, Star has the lowest CME (5.059\,m/s) and RME ($14.480^\circ$/s) among methods with the same inputs. Kalman filtering of the same MoCA4D observations gives a smaller center error (6.822 vs.\ 6.893\,m) and higher IoU (16.867\% vs.\ 16.610\%). On ADT, Star improves all four metrics over this Kalman filter and has the lowest center error in the common-input group (1.212\,m). Mean IoU remains below 1\% for every method in this group. Section~\ref{sec:supp-adt-diagnosis} examines the center errors behind the low overlap on ADT.

\paragraph{Motion-error metrics.}
For scored observations $s$ and $t$ separated by $\Delta t$ seconds, let $\mathbf{e}_t=\mathbf{Q}_t(\hat{\mathbf{c}}_t-\mathbf{c}_t^*)$ be the center error vector in world coordinates, where $\mathbf{Q}_t$ rotates vectors from camera to world coordinates. The change in this vector gives the error in center displacement, so
\begin{equation}
\mathrm{CME}_{s,t}=\frac{\|\mathbf{e}_t-\mathbf{e}_s\|_2}{\Delta t}.
\end{equation}
A track with a fixed nonzero position error in world coordinates has nonzero CE and zero CME. For RME, we apply the ground-truth change in orientation $\Delta\mathbf{R}_{s\rightarrow t}^*$ to the earlier predicted orientation $\hat{\mathbf{R}}_s$. We compare this orientation with the later prediction $\hat{\mathbf{R}}_t$ and divide their angular distance by the elapsed time, giving
\begin{equation}
\mathrm{RME}_{s,t}=\frac{d_{\mathrm{box}}(\Delta\mathbf{R}_{s\rightarrow t}^*\hat{\mathbf{R}}_s,\hat{\mathbf{R}}_t)}{\Delta t}.
\end{equation}
The distance $d_{\mathrm{box}}$ is in degrees. It matches cuboid axes by dimension ratio over the 24 proper signed permutations and breaks ties by the smallest rotation angle. Equivalent box orientations therefore receive the same score regardless of how their axes are labeled. Ground-truth motion enters only at evaluation.

\section{Object motion in dynamic-object tracking}
\label{sec:supp-dynamic-tracking}
We extend the comparison in Table~\ref{tab:real_world_tracking} to 150 HOT3D tracks and 442 HOI4D tracks (Table~\ref{tab:dynamic-counts}). We selected tracks from ground-truth motion after the initial evaluation and fixed the selection before comparing methods. A frame pair counts as moving if GT translation speed is at least $0.05$\,m/s or reliable rotation speed is at least $5^\circ$/s (Sec.~\ref{sec:supp-real-motion}).
We select tracks with at least $0.5$\,s of consecutive moving pairs. We score all eligible observations on these tracks, including those during stationary periods. Each recording contributes equally to the dataset mean. The 10 ADT recordings come from one apartment, and the HOT3D and HOI4D recordings cover two participants and one participant, respectively. Adding $B$ reduced fallbacks on AV2, ADT and HOT3D. Both Star variants completed the HOI4D tracks without a fallback.

\begin{table*}[!tbp]
\centering\small
\caption{\textbf{Dynamic-object tracks by dataset.} We count each object once per original recording and group HOT3D clips by recording. Scores include fallback predictions. Bold marks the lower fallback count in each row, including ties.}
\label{tab:dynamic-counts}
\setlength{\tabcolsep}{4pt}
\begin{tabular*}{\textwidth}{@{\extracolsep{\fill}}lcccccc@{}}
\toprule
Dataset & Recordings & Objects & Box events & Motion pairs & $A$-only fallbacks & $A+B$ fallbacks \\
\midrule
AV2 & 150 & 2,692 & 56,366 & 50,198 & 26 & \textbf{1} \\
ADT & 10 & 109 & 32,570 & 30,711 & 49 & \textbf{13} \\
HOT3D & 30 & 150 & 62,651 & 59,908 & 11 & \textbf{3} \\
HOI4D & 442 & 442 & 61,872 & 60,102 & \textbf{0} & \textbf{0} \\
\bottomrule
\end{tabular*}
\end{table*}

\paragraph{Contribution of object motion.}
On the HOT3D and HOI4D validation tracks, $A+B$ Star gives the highest IoU and the lowest errors among the four methods (Table~\ref{tab:dynamic-in-domain}). On HOT3D, rotation-motion error falls from $135.86^\circ$/s with observation-only Kalman to $68.08^\circ$/s with camera-only Star and $66.20^\circ$/s with $A+B$ Star. Most of the reduction relative to Kalman is already present when Star uses camera motion alone.

\begin{table*}[!tbp]
\centering\footnotesize
\caption{\textbf{Tracking on HOT3D and HOI4D validation tracks.} All methods use the same MoCA4D observations. Both datasets contributed training data for branch adaptation. Bold marks the best score in each column.}
\label{tab:dynamic-in-domain}
\setlength{\tabcolsep}{4pt}
\begin{tabular*}{\textwidth}{@{\extracolsep{\fill}}lcccccccc@{}}
\toprule
& \multicolumn{4}{c}{HOT3D} & \multicolumn{4}{c}{HOI4D} \\
\cmidrule(lr){2-5}\cmidrule(l){6-9}
Method & IoU $\uparrow$ & CE $\downarrow$ & CME $\downarrow$ & RME $\downarrow$ & IoU $\uparrow$ & CE $\downarrow$ & CME $\downarrow$ & RME $\downarrow$ \\
 & (\%) & (m) & (m/s) & ($^\circ$/s) & (\%) & (m) & (m/s) & ($^\circ$/s) \\
\midrule
MoCA4D observation & 23.221 & 0.073 & 0.263 & 168.253 & 33.165 & 0.062 & 0.144 & 74.241 \\
MoCA4D observation + Kalman & 20.385 & 0.077 & 0.186 & 135.858 & 33.414 & 0.061 & 0.077 & 42.221 \\
MoCA4D Star, $A$ only & 23.563 & 0.069 & 0.112 & 68.083 & 33.116 & 0.061 & 0.075 & 27.469 \\
MoCA4D Star, $A+B$ & \textbf{23.702} & \textbf{0.068} & \textbf{0.109} & \textbf{66.196} & \textbf{33.645} & \textbf{0.060} & \textbf{0.073} & \textbf{27.225} \\
\bottomrule
\end{tabular*}
\end{table*}

Table~\ref{tab:dynamic-ci} pairs $A+B$ Star with camera-only Star (upper rows) and with constant-velocity Kalman filtering of the same observations (lower rows). The two Star variants propagate their own box estimates and differ only in whether they use the predicted object transform or set $B=I$. Adding $B$ lowers rotation-motion error on HOT3D but raises it on AV2. For the box center, adding $B$ reduces both position error and frame-to-frame motion error on all four datasets.

\begin{table*}[!tbp]
\centering\footnotesize
\caption{\textbf{Paired differences on dynamic-object tracks.} Each entry is the $A+B$ Star score minus the control score. Bold estimates favor $A+B$ and underlined estimates favor the control. Brackets give individual 95\% intervals from 2,000 paired bootstrap resamples of the original recordings.}
\label{tab:dynamic-ci}
\setlength{\tabcolsep}{4pt}
\begin{tabular*}{\textwidth}{@{\extracolsep{\fill}}lcccc@{}}
\toprule
Dataset, control & $\Delta$IoU (pp) & $\Delta$CE (mm) & $\Delta$CME (m/s) & $\Delta$RME ($^\circ$/s) \\
\midrule
AV2, $A$ only & $\mathbf{+3.061}$ & $\mathbf{-387.25}$ & $\mathbf{-0.0519}$ & $\underline{+0.386}$ \\
 & {\scriptsize $[+2.514,\,+3.705]$} & {\scriptsize $[-472.71,\,-307.03]$} & {\scriptsize $[-0.0882,\,-0.0186]$} & {\scriptsize $[+0.020,\,+0.746]$} \\[2pt]
ADT, $A$ only & $\mathbf{+0.131}$ & $\mathbf{-6.82}$ & $\mathbf{-0.0018}$ & $\mathbf{-0.698}$ \\
 & {\scriptsize $[+0.046,\,+0.210]$} & {\scriptsize $[-7.49,\,-6.14]$} & {\scriptsize $[-0.0028,\,-0.0010]$} & {\scriptsize $[-1.357,\,-0.061]$} \\[2pt]
HOT3D, $A$ only & $\mathbf{+0.139}$ & $\mathbf{-0.61}$ & $\mathbf{-0.0021}$ & $\mathbf{-1.887}$ \\
 & {\scriptsize $[-0.078,\,+0.332]$} & {\scriptsize $[-1.10,\,-0.14]$} & {\scriptsize $[-0.0032,\,-0.0012]$} & {\scriptsize $[-2.634,\,-1.148]$} \\[2pt]
HOI4D, $A$ only & $\mathbf{+0.529}$ & $\mathbf{-1.08}$ & $\mathbf{-0.0015}$ & $\mathbf{-0.244}$ \\
 & {\scriptsize $[+0.337,\,+0.709]$} & {\scriptsize $[-1.42,\,-0.72]$} & {\scriptsize $[-0.0018,\,-0.0013]$} & {\scriptsize $[-0.508,\,+0.033]$} \\[2pt]
\midrule
AV2, Kalman & $\underline{-1.055}$ & $\underline{+152.97}$ & $\mathbf{-2.8277}$ & $\mathbf{-19.402}$ \\
 & {\scriptsize $[-1.556,\,-0.558]$} & {\scriptsize $[+64.64,\,+243.01]$} & {\scriptsize $[-2.9687,\,-2.6910]$} & {\scriptsize $[-22.464,\,-16.282]$} \\[2pt]
ADT, Kalman & $\mathbf{+0.584}$ & $\mathbf{-8.90}$ & $\mathbf{-0.2447}$ & $\mathbf{-43.540}$ \\
 & {\scriptsize $[+0.423,\,+0.739]$} & {\scriptsize $[-14.27,\,-2.45]$} & {\scriptsize $[-0.2802,\,-0.2109]$} & {\scriptsize $[-48.254,\,-39.169]$} \\[2pt]
HOT3D, Kalman & $\mathbf{+3.317}$ & $\mathbf{-8.85}$ & $\mathbf{-0.0762}$ & $\mathbf{-69.662}$ \\
 & {\scriptsize $[+2.448,\,+4.218]$} & {\scriptsize $[-10.41,\,-7.32]$} & {\scriptsize $[-0.0818,\,-0.0709]$} & {\scriptsize $[-77.464,\,-61.751]$} \\[2pt]
HOI4D, Kalman & $\mathbf{+0.232}$ & $\mathbf{-1.30}$ & $\mathbf{-0.0041}$ & $\mathbf{-14.995}$ \\
 & {\scriptsize $[+0.055,\,+0.415]$} & {\scriptsize $[-1.59,\,-1.02]$} & {\scriptsize $[-0.0053,\,-0.0029]$} & {\scriptsize $[-16.933,\,-13.100]$} \\[2pt]
\bottomrule
\end{tabular*}
\end{table*}

\paragraph{Track selection.}
Table~\ref{tab:dynamic-sensitivity} uses the same predictions for five definitions of a dynamic track. We vary the minimum motion duration from 0.5\,s to 0.25 or 1.0\,s, or keep it at 0.5\,s and halve or double both speed thresholds. The mean IoU favors $A+B$ under every definition, although the HOT3D intervals also admit zero gain.

\begin{table*}[!tbp]
\centering\footnotesize
\caption{\textbf{Motion thresholds for selecting dynamic objects.} Entries give the IoU difference ($A+B$ minus $A$ only) in percentage points. Bold marks the largest observed difference in each dataset. Intervals are computed as in Table~\ref{tab:dynamic-ci}.}
\label{tab:dynamic-sensitivity}
\setlength{\tabcolsep}{4pt}
\begin{tabular*}{\textwidth}{@{\extracolsep{\fill}}lcccc@{}}
\toprule
Selection & AV2 & ADT & HOT3D & HOI4D \\
\midrule
0.25 s, default speeds & $+2.827$ & $+0.132$ & $+0.144$ & $+0.528$ \\
 & {\scriptsize $[+2.256,\,+3.386]$} & {\scriptsize $[+0.056,\,+0.216]$} & {\scriptsize $[-0.082,\,+0.342]$} & {\scriptsize $[+0.346,\,+0.725]$} \\[2pt]
0.5 s, default speeds & $+3.061$ & $+0.131$ & $+0.139$ & $+0.529$ \\
 & {\scriptsize $[+2.514,\,+3.705]$} & {\scriptsize $[+0.046,\,+0.210]$} & {\scriptsize $[-0.078,\,+0.332]$} & {\scriptsize $[+0.337,\,+0.709]$} \\[2pt]
1.0 s, default speeds & $\mathbf{+3.672}$ & $\mathbf{+0.145}$ & $+0.113$ & $+0.558$ \\
 & {\scriptsize $[+3.001,\,+4.391]$} & {\scriptsize $[+0.057,\,+0.232]$} & {\scriptsize $[-0.155,\,+0.329]$} & {\scriptsize $[+0.356,\,+0.753]$} \\[2pt]
0.5 s, half speeds & $+2.807$ & $+0.133$ & $\mathbf{+0.154}$ & $+0.528$ \\
 & {\scriptsize $[+2.241,\,+3.382]$} & {\scriptsize $[+0.053,\,+0.218]$} & {\scriptsize $[-0.068,\,+0.353]$} & {\scriptsize $[+0.340,\,+0.715]$} \\[2pt]
0.5 s, double speeds & $+3.671$ & $+0.134$ & $+0.107$ & $\mathbf{+0.593}$ \\
 & {\scriptsize $[+2.965,\,+4.386]$} & {\scriptsize $[+0.050,\,+0.214]$} & {\scriptsize $[-0.167,\,+0.329]$} & {\scriptsize $[+0.386,\,+0.803]$} \\[2pt]
\bottomrule
\end{tabular*}
\end{table*}

Table~\ref{tab:cohort-distinction} scores Star over different portions of the data. On ADT, IoU is 8.127\% for moving events and 2.784\% for whole tracks of dynamic objects. ADT's own dynamic labels select 546 tracks with 135,973 box observations and give an IoU of 1.694\%. Scoring only the moments of motion gives a substantially higher estimate of ADT overlap than scoring the same tracker through stationary periods as well.

\begin{table}[!htb]
\centering\small
\caption{\textbf{Star IoU (\%) under three scoring rules.} All columns use the same predictions. Whole tracks include stationary periods. Bold marks the highest observed IoU in each row.}
\label{tab:cohort-distinction}
\setlength{\tabcolsep}{4pt}
\begin{tabular*}{\columnwidth}{@{\extracolsep{\fill}}lccc@{}}
\toprule
Dataset & All & Moving & Dynamic objects \\
 & objects & events only & whole tracks \\
\midrule
AV2 & 16.610 & 18.156 & \textbf{19.528} \\
ADT & 0.915 & \textbf{8.127} & 2.784 \\
HOT3D & 22.116 & 21.960 & \textbf{23.702} \\
HOI4D & 33.569 & \textbf{33.977} & 33.645 \\
\bottomrule
\end{tabular*}
\end{table}

\section{ADT localization diagnosis}
\label{sec:supp-adt-diagnosis}
We preserve the full ADT support of 10 recordings and 431,681 box events and replace one predicted box component at a time with GT. Table~\ref{tab:adt-oracle} reports the resulting IoU. These oracle interventions diagnose sensitivity to a component; they are not new model results, additive causal contributions, or a guaranteed upper bound. The same center-dominated pattern occurs for raw observations, Kalman, and Star.

\begin{table}[!htb]
\centering\small
\caption{\textbf{One-component GT replacement on full ADT.} IoU (\%). Each replacement keeps the other two predicted components fixed.}
\label{tab:adt-oracle}
\setlength{\tabcolsep}{2pt}
\begin{tabular*}{\columnwidth}{@{\extracolsep{\fill}}lrrrr@{}}
\toprule
Method & Original & \shortstack{GT\\center} & GT size & \shortstack{GT\\rotation} \\
\midrule
Observation & 0.932 & 24.445 & 1.476 & 0.947 \\
Obs.\ + Kalman & 0.587 & 25.255 & 1.052 & 0.600 \\
Star & 0.915 & 24.247 & 1.462 & 0.941 \\
\bottomrule
\end{tabular*}
\par\smallskip\raggedright
A box can track frame-to-frame displacement accurately even when a persistent offset in its center leaves little overlap with the ground truth.
\end{table}

For Star, center replacement improves IoU by 23.332 percentage points, versus 0.547 for size and 0.025 for rotation. Mean absolute error parallel to the GT viewing ray is 1.208\,m, versus 0.041\,m transverse error. This supports center/range error as a major restriction on overlap. It does not identify whether training, depth estimation, preprocessing, or another upstream factor causes that error.

\section{Star Optimization analyses}
\label{sec:supp-star}
\subsection{History and filtering in box refinement}

\begin{table*}[!tbp]
\centering\small
\caption{\textbf{Star and Kalman updates on dynamic-object tracks.} All four estimators use the same observations and learned camera and object motion on the tracks from Table~\ref{tab:real_world_tracking}. Bold marks the best mean in each column.}
\label{tab:history-matched}
\begin{tabular*}{\textwidth}{@{\extracolsep{\fill}}lcccccccc@{}}\toprule
 & \multicolumn{4}{c}{AV2 (2,692 objects)} & \multicolumn{4}{c}{ADT (109 objects)}\\
\cmidrule(lr){2-5}\cmidrule(l){6-9}
Estimator & IoU $\uparrow$ & CE $\downarrow$ & CME $\downarrow$ & RME $\downarrow$ & IoU $\uparrow$ & CE $\downarrow$ & CME $\downarrow$ & RME $\downarrow$\\
 & (\%) & (m) & (m/s) & ($^\circ$/s) & (\%) & (m) & (m/s) & ($^\circ$/s)\\\midrule
Star, one history edge & \textbf{20.376} & 6.289 & 4.954 & \textbf{8.079} & 2.695 & \textbf{0.611} & 0.464 & \textbf{44.833}\\
Star, three history edges & 19.528 & 6.381 & 4.826 & 12.788 & 2.784 & 0.612 & 0.449 & 67.531\\\addlinespace
Kalman & 20.293 & \textbf{6.265} & 4.891 & 13.420 & 2.721 & 0.614 & 0.469 & 72.940\\
Kalman, low $Q$ & 18.655 & 6.512 & \textbf{3.586} & 9.444 & \textbf{2.820} & 0.614 & \textbf{0.357} & 48.306\\\bottomrule
\end{tabular*}
\end{table*}

\begin{table*}[!tbp]
\centering\footnotesize
\caption{\textbf{Response to changes in object motion.} Each method is scored on the same episodes. CE is absolute center error (m), Disp. is displacement error after aligning centers at onset (m), and Time is absolute timing error (ms). Bold marks the lowest mean for each metric within a row, including ties.}
\label{tab:maneuver-summary}
\setlength{\tabcolsep}{2pt}
\begin{tabular*}{\textwidth}{@{\extracolsep{\fill}}llccccccccccccc@{}}\toprule
 & & & \multicolumn{3}{c}{Star, one edge} & \multicolumn{3}{c}{Star, three edges} & \multicolumn{3}{c}{Kalman} & \multicolumn{3}{c}{Kalman, low $Q$}\\
\cmidrule(lr){4-6}\cmidrule(lr){7-9}\cmidrule(lr){10-12}\cmidrule(l){13-15}
Dataset & Maneuver & Episodes & CE & Disp. & Time & CE & Disp. & Time & CE & Disp. & Time & CE & Disp. & Time\\\midrule
AV2 & Acceleration & 1 & \textbf{1.131} & 0.212 & \textbf{120} & 1.251 & 0.166 & \textbf{120} & \textbf{1.131} & 0.212 & \textbf{120} & 1.448 & \textbf{0.129} & \textbf{120}\\
 & Deceleration & 2 & 7.116 & 0.917 & 180 & 6.991 & \textbf{0.399} & \textbf{130} & 7.113 & 0.918 & 180 & \textbf{6.810} & 0.612 & 230\\\addlinespace
ADT & Acceleration & 4 & 0.155 & 0.041 & \textbf{75} & 0.151 & \textbf{0.038} & 80 & 0.155 & 0.041 & \textbf{75} & \textbf{0.147} & 0.039 & 85\\
 & Deceleration & 2 & 0.137 & 0.066 & \textbf{80} & 0.132 & 0.055 & \textbf{80} & 0.137 & 0.066 & \textbf{80} & \textbf{0.119} & \textbf{0.047} & \textbf{80}\\
 & Turn & 2 & \textbf{0.132} & \textbf{0.037} & 80 & 0.141 & 0.044 & \textbf{70} & \textbf{0.132} & \textbf{0.037} & 80 & 0.152 & 0.047 & \textbf{70}\\
\bottomrule
\end{tabular*}
\end{table*}

\paragraph{One or three history edges.}
We use three history edges to stabilize the depth estimate along the observed viewing ray. Each edge propagates a different past box to the current frame. With one edge, the most recent box carries all the historical weight. On AV2 and ADT, three edges lower center-motion error by about 3\% relative to one (Table~\ref{tab:history-matched}). One edge, in turn, gives 34--37\% lower rotation-motion error than three. When rotation is the priority, the same trained model can use one edge at inference.

\paragraph{Dynamics-aware Kalman update.}
Star treats a box propagated by learned motion as a second estimate of the pose seen in the current frame. Our Kalman variant uses the same idea in a recursive filter. It propagates the last corrected box and updates its depth and orientation with the current observation. On the observed ray, $z^+=z^-+K(z_{\rm obs}-z^-)$. We normalize the measurement variance to $R=1$. At regular intervals, a process variance of $Q=1/6$ gives a steady-state gain of $1/3$, matching Star's observation weight for depth. Lowering $Q$ to $1/24$ reduces the gain to $0.184$. This gives the lowest center-motion error on both datasets, although AV2 absolute center error rises from 6.265 to 6.512\,m (Table~\ref{tab:history-matched}).

\paragraph{Response during maneuvers.}
Table~\ref{tab:maneuver-summary} covers 11 nonoverlapping episodes selected from ground-truth velocity changes. We compare the times when predicted and ground-truth velocity first reach half the annotated change along its direction, capping missing responses at 0.6\,s. Three-edge Star has a smaller mean timing error than one-edge Star in the two AV2 decelerations (130 vs.\ 180\,ms). In the two ADT decelerations, all four estimators average 80\,ms.

\subsection{Weighting and feedback controls}

\begin{table*}[!tbp]
\centering\small
\caption{\textbf{Fusion alternatives on the full AV2 and ADT evaluation.} The rows share observation predictions and scored support. Uniform Star and recent Star differ in both recency weights and $\lambda_c$; their difference does not isolate recency weighting. Bold marks the best value per column.}
\label{tab:fusion-alternatives}
\begin{tabular*}{\textwidth}{@{\extracolsep{\fill}}lrrrrrrrr@{}}
\toprule
& \multicolumn{4}{c}{AV2} & \multicolumn{4}{c}{ADT} \\
Method & IoU $\uparrow$ & CE $\downarrow$ & CME $\downarrow$ & RME $\downarrow$ & IoU $\uparrow$ & CE $\downarrow$ & CME $\downarrow$ & RME $\downarrow$ \\
& (\%) & (m) & (m/s) & ($^\circ$/s) & (\%) & (m) & (m/s) & ($^\circ$/s) \\
\midrule
Observation & 17.826 & 6.909 & 14.861 & 36.139 & \textbf{0.932} & 1.274 & 2.759 & 202.176 \\
Observation + Kalman & 16.867 & 6.822 & 8.179 & 39.493 & 0.587 & 1.252 & 1.640 & 112.628 \\
Observation / dynamic 1:1 & \textbf{17.876} & 6.800 & 8.684 & 24.839 & 0.897 & 1.247 & 1.699 & 140.484 \\
Uniform Star, $\lambda_c=0.25$ & 17.212 & \textbf{6.797} & 7.462 & 14.486 & 0.912 & 1.222 & 1.437 & \textbf{74.356} \\
Chain Huber & 17.626 & 6.833 & 5.980 & 16.644 & 0.864 & 1.295 & 1.309 & 106.035 \\
Recent Star, $\lambda_c=0.5$ & 16.610 & 6.893 & \textbf{5.059} & \textbf{14.480} & 0.915 & \textbf{1.212} & \textbf{1.004} & 74.363 \\
\bottomrule
\end{tabular*}
\end{table*}

\paragraph{Fusion alternatives.}
Table~\ref{tab:fusion-alternatives} reports absolute accuracy together with motion errors for simple averaging, filtering, and motion-based fusion. Recent Star yields the lowest CME among these MoCA4D variants, while other configurations attain lower AV2 center error or higher IoU. We interpret this as a trade-off between localization and motion accuracy.

\paragraph{Parameter sweep and rollout controls.}
We analyze 48-frame windows from 110 recordings (9,180 scored box events and 8,313 motion pairs) to compare weighting and feedback policies. Absolute results use equal-weight dataset macro averaging after object and recording aggregation. The nine settings combine $\lambda_c\in\{0.125,0.25,0.5\}$ with uniform $(1,1,1)/3$, recent $(1,2,4)/7$, or latest-only $(0,0,1)$ center weights, ordered from oldest to newest. We keep $w_o=0.25$ and $\lambda_R=1$ fixed. Table~\ref{tab:controlled-star} separates two computations: fixed-candidate recombination uses the reference uniform-$\lambda_c=0.25$ candidates and rotations, whereas each recursive rollout reuses its own corrected outputs.

\begin{table*}[!tbp]
\centering\small
\caption{\textbf{Star center-weight sweep: all nine settings.} Fixed-candidate columns diagnose center recombination; recursive columns report actual inference trajectories. Bold marks the best displayed value among the three policies at each fixed $\lambda_c$. Rotation changes are small and should not be interpreted as substantial gains.}
\label{tab:controlled-star}
\setlength{\tabcolsep}{4pt}
\renewcommand{\arraystretch}{1.10}
\begin{tabular*}{\textwidth}{@{\extracolsep{\fill}}lrrrrrrr@{}}
\toprule
& & \multicolumn{2}{c}{Fixed candidates} & \multicolumn{4}{c}{Recursive rollout} \\
\cmidrule(lr){3-4}\cmidrule(lr){5-8}
Policy & $\lambda_c$ & CE $\downarrow$ & CME $\downarrow$ & IoU $\uparrow$ & CE $\downarrow$ & CME $\downarrow$ & RME $\downarrow$ \\
& & (m) & (m/s) & (\%) & (m) & (m/s) & ($^\circ$/s) \\
\midrule
Uniform & 0.125 & \textbf{2.0459} & 3.3470 & \textbf{26.9751} & \textbf{2.0445} & 3.3158 & \textbf{41.1698} \\
Recent & 0.125 & 2.0492 & \textbf{3.2860} & 26.9652 & 2.0500 & 3.2328 & 41.1974 \\
Latest only & 0.125 & 2.0610 & 3.3054 & 26.8770 & 2.0691 & \textbf{3.2051} & 41.1975 \\
\midrule
Uniform & 0.25 & \textbf{2.0352} & 2.5406 & 27.0980 & \textbf{2.0352} & 2.5406 & 41.2063 \\
Recent & 0.25 & 2.0357 & \textbf{2.4269} & \textbf{27.1068} & 2.0355 & \textbf{2.4285} & 41.2064 \\
Latest only & 0.25 & 2.0494 & 2.6329 & 26.9957 & 2.0562 & 2.4531 & \textbf{41.1883} \\
\midrule
Uniform & 0.5 & 2.0549 & 1.8615 & 26.9324 & 2.0814 & 1.8638 & 41.2061 \\
Recent & 0.5 & \textbf{2.0466} & \textbf{1.7155} & 27.0675 & \textbf{2.0654} & \textbf{1.7185} & \textbf{41.2059} \\
Latest only & 0.5 & 2.0589 & 2.2814 & \textbf{27.0929} & 2.0724 & 1.7886 & 41.2062 \\
\bottomrule
\end{tabular*}
\end{table*}

\begin{table*}[!tbp]
\centering\small
\caption{\textbf{History feedback at fixed Star parameters.} Bold marks the best displayed value per column. Recursive feedback has the lowest RME, while absolute center accuracy and IoU favor other policies. All outputs, including fallbacks, enter the aggregation; the observed fallback rate is zero for these paths.}
\label{tab:star-feedback}
\setlength{\tabcolsep}{6pt}
\renewcommand{\arraystretch}{1.10}
\begin{tabular*}{\textwidth}{@{\extracolsep{\fill}}lrrrrr@{}}
\toprule
History policy & IoU (\%) $\uparrow$ & CE (m) $\downarrow$ & CME (m/s) $\downarrow$ & Box RE ($^\circ$) $\downarrow$ & RME ($^\circ$/s) $\downarrow$ \\
\midrule
Raw-observation seeds & 27.1237 & \textbf{2.0280} & \textbf{2.5181} & 32.8529 & 64.6550 \\
Nonrecursive correction & \textbf{27.1570} & 2.0309 & 2.5470 & 32.6426 & 56.0739 \\
Recursive feedback & 27.0980 & 2.0352 & 2.5406 & \textbf{32.6123} & \textbf{41.2063} \\
\bottomrule
\end{tabular*}
\end{table*}

At $\lambda_c=0.5$, recent weights give lower recursive CME than uniform or latest-only weights. With recent weights, raising $\lambda_c$ from 0.25 to 0.5 lowers CME by 0.7100\,m/s but increases CE by 0.0299\,m. The IoU change is $-0.0393$ percentage points. These are differences of the reported absolute macro averages. The fixed-candidate diagnostic preserves the reference state path and cannot substitute for a recursive rollout when reporting configuration performance.

\paragraph{Feedback policy.}
Table~\ref{tab:star-feedback} holds uniform center weights, $\lambda_c=0.25$, $\lambda_R=1$, and $w_o=0.25$ fixed on the same 110 recordings. A seed is the box pose that Star propagates to later frames. The raw-observation baseline uses the center and orientation predicted by the observation branch as its seed. The nonrecursive control uses the pose refined by that baseline at the same frame. With recursive feedback, the tracker uses its own refined pose. This comparison tests the history update policy at a fixed fusion setting.

Recursive feedback lowers RME by 23.4487$^\circ$/s relative to raw-observation seeds and by 14.8676$^\circ$/s relative to nonrecursive correction. Its CE is $0.0072$\,m higher than that of the raw-observation baseline, and its IoU is $0.0590$ percentage points lower than that of the nonrecursive control. These matched-window comparisons support a rotation-motion benefit, with small changes in absolute box accuracy.

\end{document}